\documentclass[pdflatex,sn-mathphys-num]{sn-jnl}

\usepackage{graphicx}%
\usepackage{multirow}%
\usepackage{amsmath,amssymb,amsfonts}%
\usepackage{amsthm}%
\usepackage{mathrsfs}%
\usepackage[title]{appendix}%
\usepackage{xcolor}%
\usepackage{textcomp}%
\usepackage{manyfoot}%
\usepackage{booktabs}%
\usepackage{algorithm}%
\usepackage{algorithmicx}%
\usepackage{algpseudocode}%
\usepackage{listings}%
\usepackage{longtable}
\usepackage{soul}
\usepackage{ulem}
\usepackage{array}
\usepackage{hhline}
\usepackage{fancyhdr} 
\AtBeginDocument{%
    \thispagestyle{fancy}%
    \fancyhf{}%
    \fancyhead[C]{Accepted by Journal of Imaging Informatics in Medicine}%
}

\theoremstyle{thmstyleone}%
\theoremstyle{thmstyletwo}%

\theoremstyle{thmstylethree}%

\begin{document}

\title[MHNet: Multi-view High-order Network for Diagnosing Neurodevelopmental
Disorders Using Resting-state fMRI]{MHNet: Multi-view High-order Network for Diagnosing Neurodevelopmental
Disorders Using Resting-state fMRI}


\author[1]{\fnm{Yueyang} \sur{Li}}\email{lyy20010615@163.com}
\author*[1]{\fnm{Weiming} \sur{Zeng}}\email{zengwm86@163.com}
\author[1]{\fnm{Wenhao} \sur{Dong}}\email{jsxzdwh@163.com}
\author[1]{\fnm{Luhui} \sur{Cai}}\email{clh0x123@gmail.com}
\author[1]{\fnm{Lei} \sur{Wang}}\email{wxlluck@gmail.com}
\author[1]{\fnm{Hongyu} \sur{Chen}}\email{hongychen676@gmail.com}
\author[2]{\fnm{Hongjie} \sur{Yan}}\email{yanhjns@gmail.com}
\author[3]{\fnm{Lingbin} \sur{Bian}}\email{bianlb@shanghaitech.edu.cn}
\author*[4]{\fnm{Nizhuan} \sur{Wang}}\email{wangnizhuan1120@gmail.com}

\affil*[1]{\orgdiv{Lab of Digital Image and Intelligent Computation}, \orgname{Shanghai Maritime University}, \orgaddress{\city{Shanghai}, \postcode{201306}, \country{China}}}
\affil[2]{\orgdiv{Department of Neurology}, \orgname{Affiliated Lianyungang Hospital of Xuzhou Medical University}, \orgaddress{\city{Lianyungang}, \postcode{222002}, \country{China}}}
\affil[3]{\orgdiv{School of Biomedical Engineering \& State Key Laboratory of Advanced Medical Materials and Devices}, \orgname{ShanghaiTech University}, \orgaddress{\city{Shanghai}, \postcode{201210}, \country{China}}}
\affil*[4]{\orgdiv{Department of Chinese and Bilingual Studies}, \orgname{The Hong Kong Polytechnic University}, \orgaddress{\city{Hung Hom, Kowloon}, \country{Hong Kong Special Administrative Region, China}}}

\abstract{\textbf{Background:} Deep learning models have shown promise in diagnosing neurodevelopmental disorders (NDD) like ASD and ADHD. However, many models either use graph neural networks (GNN) to construct single-level brain functional networks (BFNs) or employ spatial convolution filtering for local information extraction from rs-fMRI data, often neglecting high-order features crucial for NDD classification.
\textbf{Methods:} We introduce a Multi-view High-order Network (MHNet) to capture hierarchical and high-order features from multi-view BFNs derived from rs-fMRI data for NDD prediction. MHNet has two branches: the Euclidean Space Features Extraction (ESFE) module and the Non-Euclidean Space Features Extraction (Non-ESFE) module, followed by a Feature Fusion-based Classification (FFC) module for NDD identification. ESFE includes a Functional Connectivity Generation (FCG) module and a High-order Convolutional Neural Network (HCNN) module to extract local and high-order features from BFNs in Euclidean space. Non-ESFE comprises a Generic Internet-like Brain Hierarchical Network Generation (G-IBHN-G) module and a High-order Graph Neural Network (HGNN) module to capture topological and high-order features in non-Euclidean space.  
\textbf{Results:} Experiments on three public datasets show that MHNet outperforms state-of-the-art methods using both AAL1 and Brainnetome Atlas templates. Extensive ablation studies confirm the superiority of MHNet and the effectiveness of using multi-view fMRI information and high-order features. Our study also offers atlas options for constructing more sophisticated hierarchical networks and explains the association between key brain regions and NDD.
\textbf{Conclusion:} MHNet leverages multi-view feature learning from both Euclidean and non-Euclidean spaces, incorporating high-order information from BFNs to enhance NDD classification performance.}




\maketitle

\section{Introduction}\label{sec1}

Autism spectrum disorder (ASD) and attention deficit hyperactivity disorder (ADHD) are two typical neurodevelopmental disorders (NDD) facing specific or sometimes overlapping challenges. Individuals with ASD often exhibit early-onset difficulty in communication and reciprocal social interactions alongside the repetitive and restricted sensory-motor behaviors \cite{ref-journal7}. On the other hand, individuals with ADHD often show persistent patterns of inattention, hyperactivity, and impulsivity that interfere with daily functioning or development \cite{ref-journal9}. 

Currently, the majority of clinical diagnoses of NDD mainly rely on the subjective assessment of abnormal behavior by clinical experts, which may lead to limited accessibility and inconsistent or delayed diagnosis. This highlights the need for more objective computer aided imaging diagnostic methods \cite{ref-journal3}. As a non-invasive imaging modality, functional magnetic resonance imaging (fMRI) has been adopted for the diagnosis of NDD in previous research studies \cite{ref-journal2}. Decoding fMRI data using machine or deep learning can provide valuable insights into the abnormality related to NDD including altered neural activity and brain connectivity, by which the disease-specific changes within the brain can be revealed \cite{ref-journal4,ref-journal6}.

Abnormalities or dysfunctions in large-scale brain functional networks (BFNs) can be reflected by altered functional connectivity (FC) derived from resting-state fMRI (rs-fMRI) data \cite{ref-journal11}. Numerous studies have discovered that brain functional impairment of NDD patients is related to abnormal FC between resting-state networks (RSNs) \cite{ref-journal12,ref-journal15}. For instance, some studies have found significant differences in the connectivity between the default mode network (DMN) and other brain regions in patients with NDD, which may be associated with the deficits in social behavior and cognitive functions \cite{ref-journal2}. However, most existing studies only focus on whole brain functional network (BFN) analysis and do not consider the hierarchical structure of the spatial topology of BFNs. This whole BFN analysis does not fully capture the complex changes in brain function of NDD patients. BFNs exhibit significant abnormalities not only in whole BFN but also in local sub-networks, which require multi-scale analytical approaches \cite{ref-journal17} for constructing networks with hierarchical structure.

Encoding both the Euclidean and non-Euclidean space features based on deep learning can reveal the complementary complex information of the BFNs. The convolutional neural networks (CNNs) can automatically extract the Euclidean space features of BFNs \cite{ref-journal2,ref-journal19}, which requires the involvement of local receptive fields encoding the values of connectivity weights of FC. The graph neural networks (GNNs) \cite{ref-journal21} has emerged as an attractive framework for modeling non-Euclidean space features of BFNs due to their powerful topological graph embedding capabilities \cite{ref-journal66}. The integration of CNNs and GNNs allows for comprehensive multi-scale feature extraction, enhancing the ability of deep learning model to generalize across different individuals and improve the NDD prediction performance.

However, simply applying CNNs or GNNs only utilizes the first-order features and cannot capture the high-order information in BFNs \cite{ref-journal22,ref-journal23}, which limits the ability of generalization and the accuracy of NDD prediction of the deep learning models. The first-order features are represented by embeddings of individual brain regions extracted from fMRI. In contrast, high-order features are crucial for revealing the complex relationships and indirect connections of the BFNs, encapsulating the intricate correlational patterns among pairwise brain regions, elucidating the topological architecture of neural networks while capturing the multifaceted interactions and emergent global properties inherent in the brain's organizational framework \cite{ref-journal68}. Capturing the high-order information of BFNs can potentially improve the generalization ability of the model.

In this paper, we proposed a novel multi-view high-order network (MHNet), consisting of the Euclidean space features extraction (ESFE) module and non-Euclidean space features extraction (Non-ESFE) module, followed by the feature fusion-based classification (FFC) module. 

(i) The ESFE module is designed to extract connectivity weights information of FC from rs-fMRI data in the Euclidean space, which contains two sub-modules, namely FC generation (FCG) module and HCNN (high-order CNN) module. FCG constructs the FC matrices using different brain atlases and maps the FC to Euclidean space. HCNN captures the Euclidean space features in FC through 1D-CNN. Specifically, the ESFE utilizes the CNN to learn the regular FC information and uses the high-order pooling (HOP) operator \cite{HOP} to formulate the discriminative and representative local features of BFNs. 

(ii) The non-ESFE module contains two sub-modules, namely generic internet-like brain hierarchical network generation (G-IBHN-G) module and HGNN (high-order GNN). To construct a comprehensive hierarchical structure of BFNs, we propose the G-IBHN-G module, consisting of Brain-WAN, Brain-MAN, and Brain-LAN components. The G-IBHN-G module encompasses all multi-scale hierarchical BFNs. This module combines automatic anatomical labeling 1 (AAL1), Brainnetome atlas, and Yeo's 7-network parcellation \cite{ref-journal24,ref-journal25}. This strategy generates multi-scale hierarchical BFNs from rs-fMRI data \cite{ref-journal26}, which is inspired by the internet classified brain hierarchical network (IBHN). Then, the non-ESFE uses residual Chebyshev Networks (ChebNet) model to capture the first-order graph features within the brain hierarchical network and uses the graph high-order pooling (GHOP) operator \cite{ref-journal27} to capture high-order topological features.

(iii) Finally, the high-order features from ESFE and non-ESFE modules are further fused in feature fusion-based classification (FFC) module to learn complementary multi-view information for NDD classification.

\section{Related Work}
This section systematically reviews the relevant works on NDD diagnosis based on fMRI data, examining methods from traditional machine learning to deep learning.

\subsection{Traditional Machine Learning-based NDD Diagnosis}
Traditional machine learning methods have been extensively applied in the diagnosis of NDD using fMRI data. Commonly used algorithms include support vector machine (SVM), random forests (RFs), and k-Nearest neighbors (kNN) \cite{ref-journal29}. SVM has been widely used due to its effectiveness in binary classification tasks such as distinguishing the brain patterns of NDD and health control \cite{ ref-journal31}. RFs and kNN offer robustness and flexibility in handling noisy and complex data. They are employed to classify ADHD patients using rs-fMRI data, showing promising accuracy and interpretability \cite{ref-journal32,knn}. However, these traditional methods often face computational inefficiencies and limited model generalization capabilities when dealing with high-dimensional and complex fMRI data \cite{ref-journal2}. Despite improvements from techniques like recursive feature elimination and principal component analysis \cite{rfe}, traditional machine learning methods remain heavily dependent on feature engineering quality and domain expertise, while being time-intensive when handling the complex heterogeneous data of NDD. These limitations necessitate the development of more advanced techniques, such as GNNs and CNNs, which offer enhanced feature learning capabilities and better classification performance to address the shortcomings of traditional methods \cite{ref-journal36}.

\subsection{GNN-based NDD Diagnosis}
GNNs are powerful tools for modeling the topology of BFNs by capturing the complex interactions between brain regions. For instance, park et al. \cite{ref-journal34} proposed a deep learning model that utilizes a residual graph convolution network (GCN) with spatio-temporal features extracted from 4D fMRI to improve the classification accuracy of ASD. Jiang et al. \cite{ref-journal36} introduced a hierarchical GCN for learning graph embeddings from brain networks to predict brain disorders. A significant advantage of GNNs is their ability to provide interpretable models. By analyzing the learned node features and edge weights, researchers can gain insights into the encoded information of specific brain regions and connections. Li et al. \cite{ref-journal37} proposed the BrainGNN which is an interpretable GNN framework to analyze fMRI and discover neurological biomarkers. In addition, GNNs are capable of encoding nuanced alterations of brain connectivity related to NDD according to the study from \cite{ref-journal38}. Yang et al. \cite{ref-journal39} proposed a new method named Pearson's correlation-based spatial constraints representation to estimate the BFNs. The development of advanced GNN architectures continues to enhance their applicability and performance, making them a promising tool in the field of NDD diagnosis. {While initial GNN applications focused on single modality analysis, recent advances have expanded to leverage multimodal data. The self-attention multi-modal triplet networks have been proposed to synergistically integrate Diffusion Tensor Imaging (DTI) and fMRI modalities to enhance diagnostic precision \cite{zhu2022multimodal}. HC-GNN employs hypercomplex neural architectures to facilitate multi-modal representation fusion while preserving model interpretability \cite{yang2024hypercomplex}. Furthermore, FC-HGNN implements a dual-stage heterogeneous graph neural framework that incorporates functional connectivity and phenotypic features to facilitate robust biomarker identification \cite{gu2025fc}.}

Recent studies have explored advanced GNNs architectures for fMRI analysis, introducing the spatio-temporal hybrid attentive graph network (ST-HAG) \cite{liu2024spatio} that leverages Transformer-based attention for temporal modeling, followed by the explainable spatio-temporal graph convolutional network (ESTGCN) \cite{chen2025explainable} for enhanced interpretability in brain network analysis. Further developments include the explainability-generalizable graph neural network (XG-GNN) \cite{qiu2024towards}, which employs meta-learning to improve model generalization, and the causality-inspired graph neural network (CI-GNN) \cite{zheng2024ci} that focuses on identifying diagnostic-relevant subgraphs. However, significant challenges remain as current models struggle with capturing hierarchical features from multi-view brain functional networks and suffer from over-smoothing issues when deepening network layers \cite{over}. Additionally, the limited representation ability of shallow GNN structures and inadequate integration of Euclidean and non-Euclidean features from multi-view data restrict their capacity to fully capture dynamic brain connectivity \cite{ref-journal40}. Therefore, we proposed a multi-scale residual GNN module within Non-ESFE that captures hierarchical and contextual graph information while addressing limitations in multi-view brain functional networks processing, over-smoothing prevention, and representation capacity enhancement for robust dynamic brain connectivity analysis.

\subsection{CNN-based NDD Diagnosis}
CNNs are particularly effective in capturing fine-grained spatial patterns in Euclidean space with local receptive fields \cite{ref-journal61}. Kawaharaa et al. \cite{ref-journal19} proposed the BrainNetCNN to predict NDD from structural brain networks of infants, and introduced a special structure with edge-to-edge, edge-to-node, and node-to-graph convolutional layers which can exploit the topological locality of brain networks. Silva et al. \cite{ref-journal62} developed a CNN-based feature extraction method combining seed correlation, local consistency, and low-frequency amplitude scores for ADHD diagnosis. {Multimodal CNN-based approaches have shown promise in neuroimaging analysis. For instance, the integration of multiple MRI modalities through CNN architectures has demonstrated remarkable potential in brain disorder diagnosis \cite{wang2018novel}. Additionally, Mao et al. \cite{ref-journal2} designed a 4D CNN architecture that incorporates multiple granularity computing and fusion models, including feature pooling, LSTM, and spatiotemporal convolution to extract dynamic features from rs-fMRI time series. Similarly, Zou et al. \cite{zou20173d} proposed a multi-modality 3D CNN architecture that combines fMRI and sMRI features for ADHD diagnosis, highlighting the advantages of brain functional and structural information fusion.}                                                                                                                  

Recent CNN-based models for neuroimaging analysis, such as local-global coupled network (LCGNet) \cite{zhou2024lcgnet} with its dual CNN-transformer backbone and static-dynamic convolutional neural network with dual pathways, have made progress in combining local and global feature extraction \cite{huang}. However, these models' inherent limitation lies in their local receptive fields, which constrain their ability to capture complex, non-Euclidean spatial information and global connectivity patterns of brain networks that extend beyond local neighborhoods. To overcome these limitations, our approach integrates CNNs with GNNs, effectively combining the strengths of both models. This fusion allows us to capture the global topological structures within the brain's network, providing a more comprehensive understanding of neural dynamics. By leveraging the non-Euclidean capabilities of GNNs alongside the local feature extraction of CNNs, our model addresses the gaps in prior studies.

\section{Method}

\subsection{Overview of MHNet}
\begin{figure}
\centering
\includegraphics[width=1\textwidth]{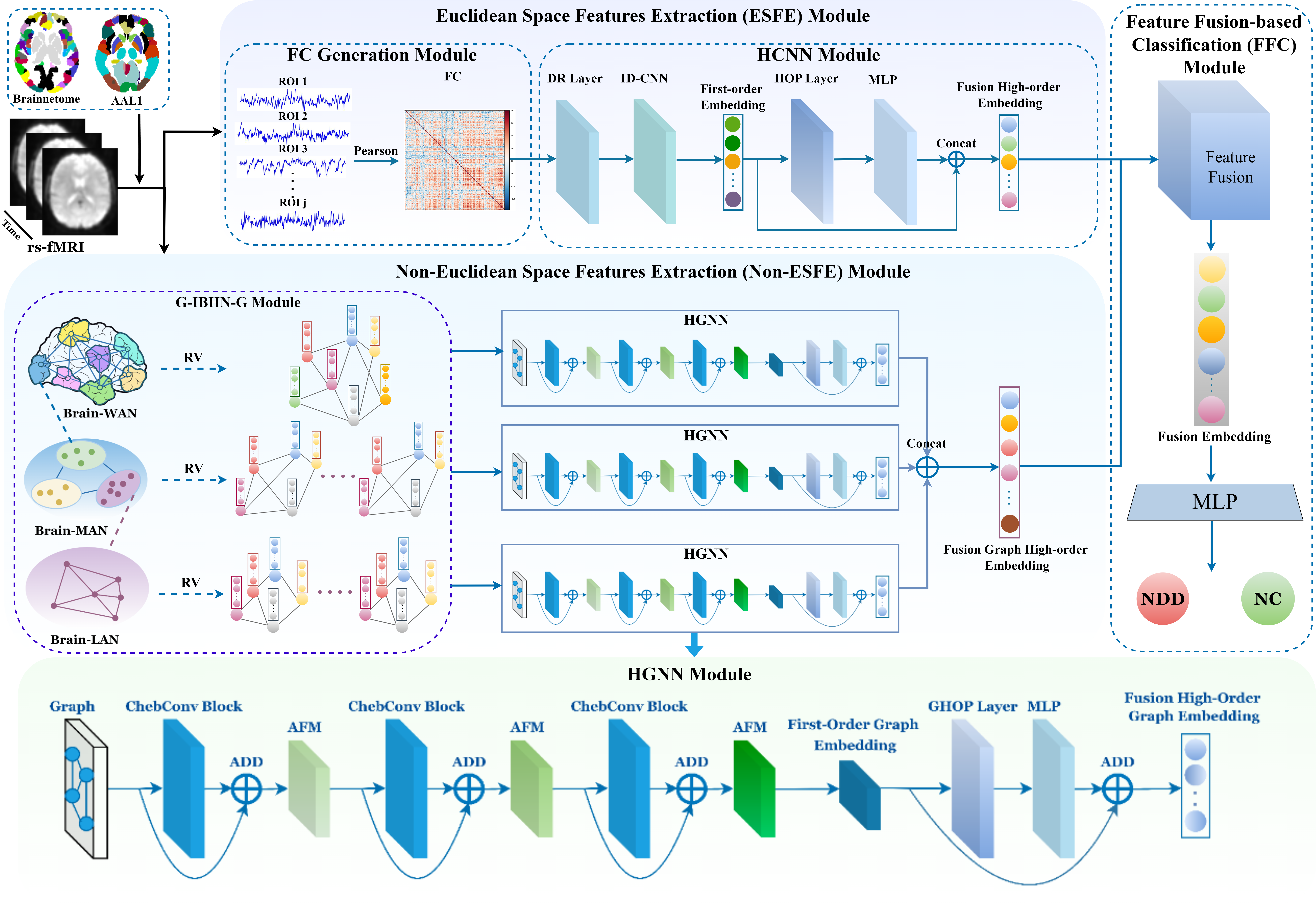}
\caption{The framework of the proposed MHNet. Multi-view data of rs-fMRI are used for the diagnosis of NDD.} \label{fig_1}
\end{figure}

As shown in Figure~\ref{fig_1}, the preprocessed rs-fMRI data are transformed into multi-view data through FC generation module and G-IBHN-G module. The transformed data comprise four different views: FC matrix, brain wide area network (Brain-WAN), brain metropolitan area network (Brain-MAN), and brain local area network (Brain-LAN). The Brain-WAN view primarily focuses on the functional connectivity and node features of the entire brain. The Brain-MAN view emphasizes the connectivity and nodel features of several sub-networks, and the Brain-LAN view delves into the local FC within a sub-network. This multi-view data representation characterizes the complex information of BFN with perspectives from whole brain to a single region. The ESFE module is dedicated to extracting high-order features of the brain in Euclidean space. The FC matrix is the input of HCNN, and the first-order embeddings, high-order embeddings, and syncretic high-order features are the output. The Non-ESFE module is dedicated to extracting high-order graph features of the brain in non-Euclidean space. The graph structures constructed by Brain-WAN, Brain-MAN and Brain-LAN are used in parallel through three HGNNs to obtain the corresponding first-order graph embedding, high-order graph embedding, and syncretic high-order graph features. Finally, the syncretic high-order features and syncretic high-order graph features are fused through the FFC module to obtain the integrated high-order features which contain the multi-view complementary information of BFNs for NDD diagnosis.

\subsection{Non-ESFE}
\subsubsection{G-IBHN-G}
To encode the hierarchical structure of the BFNs, we introduce the G-IBHN-G module which encompasses the components of Brain-WAN, Brain-MAN, and Brain-LAN. At the highest level, Brain-WAN encodes the whole BFN consisting the sub-networks of visual network (VN), somatomotor network (SMN), dorsal attention network (DAN), ventral attention network (VAN), limbic network (LN), frontoparietal network (FPN), and DMN. These sub-networks are derived from the Yeo's 7-network parcellation \cite{ref-journal25}. At the intermediate level, Brain-MAN encodes each sub-network. Taking the DMN as an example, it can be divided into finer sub-networks, such as the frontal lobe (FL) and temporal lobe (TL). At the lowest level, Brain-LAN encodes the regions (nodes) within a finer sub-network. For instance, considering the TL, it can be divided into regions including SFG\_L\_7\_2, SFG\_L\_7\_3, MFG\_L\_7\_5 based on the Brainnetome Atlas \cite{ref-journal24}, or Temporal\_Sup\_L, Temporal\_Sup\_R, Temporal\_Mid\_L, Temporal\_Mid\_L, Temporal\_Inf\_L based on AAL1.

At each level of the analysis in G-IBHN-G module, the strength of FC between sub-networks (or regions) at the same level is estimated by calculating a RV coefficient \cite{rv} defined as follows:
\begin{equation}
    {RV(A,B)}=\frac{{Tr}(AA^{\prime}BB^{\prime})}{\sqrt{T
    r\bigl[(AA^{\prime})^2\bigr]Tr\bigl[(BB^{\prime})^2\bigr]}},
\end{equation}
where \textit{A} and \textit{B} are \textit{n} $\times$ \textit{p} and \textit{n} $\times$ \textit{q} matrices representing two brain regions, $n$ is the number of samples in the rs-fMRI time series, $p$ and $q$ are the numbers of voxels in the regions of \textit{A} and \textit{B} respectively. $A^{\prime}$ and $B^{\prime}$ are the transpose of matrix $A$ and $B$ respectively, $Tr(A)$ is the trace of the matrix $A$.

\subsubsection{HGNN}
Shallow GNNs have limitation in representing complex network and long-range dependencies, as they typically only capture local node features and neighborhood information \cite{long}. This limitation hinders performance on tasks with diverse FC information for brain disease classification. Although deeper GNNs better capture intricate network structures, they often suffer from over-smoothing, resulting in indistinguishable node features and degrading model performance \cite{ ref-journal67}. The combination of Chebyshev filters and residual connections in ChebNet helps mitigate the over-smoothing problem, which preserves part of original input information at each layer and maintains the diversity of node features across layers \cite{ref-journal42}.  This strategy ensures that the model retains unique characteristic of each node while still benefits from the depth of network.

The higher-order features can encode more complex relationships and interactions between ROIs. Certain complex interactions in the brain can be easily captured by higher-order correlations, and can exhibit more complex relationships that cannot be directly modeled in lower-order statistics \cite{guo2021brain}. In the context of GNNs, high-order features extend beyond the immediate neighborhood relationships to capture sophisticated structural and semantic patterns in graph data \cite{xie2021bagfn}. To extract these high-order features, we employ a transpose operation that computes pairwise interactions between node embeddings in the latent space. Through this operation, each element in the resulting matrix represents the relationship between a pair of nodes, thereby encoding comprehensive node-to-node interactions across the entire graph \cite{ref-journal27}. This high-order representation preserves the structural information learned by previous GNN layers while simultaneously capturing more complex topological patterns that may not be directly observable in the original graph structure.

Our HGNN is able to hierarchically learn the multi-view high-order topological features in non-Euclidean space. It comprises ChebConv blocks, adaptive feature maps (AFM), GHOP layer, and multilayer perceptron (MLP). Within each ChebConv block, input features pass sequentially through the ChebConv layer, followed by a batch normalization layer, ReLU activation layer, and dropout layer. Each layer within the ChebConv block accepts node features from the preceding layer and produces updated node features for the subsequent layer. AFM effectively harnesses the features from each ChebConv block to derive multi-scale node feature embeddings.

Spectral-based graph convolution combines the overall graph structure with its individual components using the Chebyshev spectral graph convolution operator \cite{ref-journal43}. This method defines the convolution of a signal $h\in\mathbb{R}^m$ (where each node has a scalar value) with a filter $g_\theta=\mathrm{diag}(\theta)$, parameterized by $\theta\in\mathbb{R}^m$:
\begin{equation}
    g_\theta* h=Ug_\theta(\Lambda)U^\top h,
\end{equation}
where  $*$ is the convolution operator on graph. The matrix $U$, composed of the eigenvectors of the Laplace matrix $L = I-D^{-\frac{1}{2}}AD^{-\frac{1}{2}} = {U\Lambda U}^{T}$, diagonalizes $L$ as ${U\Lambda U}^{T}$, where $I$ denotes the identity matrix, $\Lambda$ represents the diagonal matrix containing the eigenvalues of $L$, and $D$ signifies the degree matrix derived from the adjacency matrix $A$ of the graph. To alleviate the computational complexity associated with computing $Ug_\theta(\Lambda)U^\top$, we approximate $g_\theta(\Lambda)$ using K-order Chebyshev polynomials. expressed as:
\begin{equation}
    g_\theta* h\approx\sum_{k=0}^{K-1}\theta_kT_k(\tilde{L})h,
\end{equation}

We define the hierarchically structured brain graphs as $G_{WAN} = (V_{w},A_{w})$, $G_{MAN} = (V_{m},A_{m})$, $G_{LAN} = (V_{l},A_{l})$ where each node represents a sub-graph (or region at the lowest level analysis in G-IBHN-G module) of the corresponding network, i.e. $V_{w} = (v_{1},\ldots,v_{n})$. In the construction of these graphs, each node encapsulates specific brain regions or functional units: nodes in $G_{WAN}$ represent major brain systems (e.g., DMN, LN), nodes in $G_{MAN}$ represent intermediate-level brain regions (e.g., FL, TL), and nodes in $G_{LAN}$ represent local brain regions defined by anatomical parcellation. The node features in $G_{WAN} = (V_{w},A_{w})$ can be expressed as the matrix $H_{w} = [x_{1}, x_{2}, \ldots, x_{n}]$, where $x_{n}$ is feature expression associated with $v_{n}$. The edges in these graphs, represented by adjacency matrices $A_w$, $A_m$, and $A_l$, are constructed based on both structural and functional connectivity. The edge weights $w_{ij}$ between nodes $i$ and $j$ are determined by:


\begin{equation}
	A_k(i,j)=\begin{cases}
		RV_{ij} & \text{if } \rho_{ij} > \gamma \\
		1 & \text{if } i = j \\
		0 & \text{otherwise} ,
	\end{cases}
\end{equation}
where $k \in {w,m,l}$ represents the network level, $\rho_{ij}$ is the RV correlation coefficient between the BOLD signals of regions $i$ and $j$, and $\gamma$ is a threshold parameter to control the graph sparsity.

For graphs $G_{MAN}$ and $G_{LAN}$, we partition the brain regions into subgraphs based on functional networks or structural partitions. These subgraphs are arranged along the diagonal of a larger sparse adjacency matrix, forming a block-diagonal matrix:

\begin{equation}
	A_k=\operatorname{diag}\left(A_{1}, A_{2}, \ldots, A_{n}\right),
\end{equation}
where each $A_i$ is the adjacency matrix of the $i$-th subgraph, $k \in {m,l}$. This configuration enables us to process all subgraphs collectively through GNN layers while preserving their individual topological characteristics.

This technique uses the graph's Laplacian eigenbasis to perform convolutions in the frequency domain, capturing both local and global information.
In this work, we define the output features of the $l-th$ ChebConv as:
\begin{equation}
    {H_{w}}^{(l+1)} = \sum_{k=0}^{K-1} \theta_k^{(l)} T_k(\tilde{L}) {H_{w}}^{(l)},
\end{equation}
where $\tilde{L}$ is the rescaled graph Laplacian, $\theta_{k}$ are the trainable parameters, $\textit{T}_k(\tilde{ {L}})=2\tilde{ {L}}T_{k-1}(\tilde{ {L}})-T_{k-2}(\tilde{ {L}})$ with $\textit{T}_k(\tilde{ {L}})=1$ and $\textit{T}_1(\tilde{ {L}})=0$. The final multi-scale graph embedding $\textit{Z}_{w}$ is obtained by aggregating the graph embeddings of all ChebConv blocks using AFM, expressed as:
\begin{equation}
    Z_w = \sum_ls^{(l)}\odot{H_w}^{(l)},
\end{equation}
where $s^{(l)}(l \in \{0,1,2\})$ represents the trainable weight following a softmax distribution, defined as:
\begin{equation}
    s^{(l)} = \text{Softmax}(r^{(l)}) = \frac{\exp(r^{(l)})}{\sum_l\exp(r^{(l)})},
\end{equation}
where $r^{\left(l\right)}$ are the learnable weights with random initalization. The learnable weights are implemented through learnable parameter matrices. AFM introduces an adaptive weighting mechanism that dynamically adjusts the weight of each layer of features through a learnable parameter matrix, so that the feature contribution of each node at different scales can be adaptively adjusted, which effectively avoids the problem of over-smoothing of features.

In the non-Euclidean space of BFN, GHOP captures high-order statistical information, which enables more complex and nuanced feature representations compared to traditional first-order pooling methods. The high-order features encapsulate both direct and indirect interactions within the BFN, thus enrich the features with high-order interactions. Taking the Brain-LAN in G-IBHN-G as an example, the GHOP scheme is used to extract high-order graph features. The high-order graph features $Z_{GHOP}$ can be expressed as:
\begin{equation}
    Z_{GHOP}=Z_l^TZ_l,
\end{equation}
where $Z_l$ represents first-order features of Brain-LAN, and $Z_l^T$ represents the transpose of first-order features. The expression of the  high-order features is:
\begin{equation}
    \tilde{Z}_l = Concat(Z_l, f_{MLP}\big(Z_{GHOP}\big)).
\end{equation}
Similarly, the high-order features $\tilde{Z}_w$ and $\tilde{Z}_m$ of brain graphs $G_{WAN} = (V_{w}, A_{w})$ and $G_{MAN} = (V_{m}, A_{m})$ can also be obtained  through the above steps.

Finally, the high-order graph features obtained through each layer of the brain network are fused to obtain the high-order graph fused features $\tilde{Z}_{GHOP}$ in non-Euclidean space:
\begin{equation}
    \tilde{Z}_{GHOP} = Concat(\tilde{Z}_w, \tilde{Z}_m, \tilde{Z}_l),
\end{equation}
where $\tilde{Z}_w$, $\tilde{Z}_m$, and $\tilde{Z}_l$ represent the high-order features obtained from Brain-WAN, Brain-MAN, and Brain-LAN respectively.

\subsection{ESFE}
The aim of our ESFE is to extract discriminative features from FC matrix apart from the features encoded from the non-ESFE. ESFE module contains the components of FCG module and HCNN module. HCNN comprises a dimensionality reduction (DR) layer, 1D-CNN layer, high-order pooling (HOP) Layer, and MLP. In ESFE module, the FC matrix is denoted as $C{\in}\mathbb{R}^{N\times N}$. The DR layer extracts the upper triangular of the FC matrix and concatenates the element values of each row in row order to form a one-dimensional feature. Subsequently, this one-dimensional feature is input into two layers of 1D-CNN and MLP, resulting in the first-order features $Z_{fc}$. 1D-CNN can effectively extract local features and capture local correlations between brain regions. Weight sharing reduces the number of parameters and enhances the robustness of the model. Through multi-layer convolution, 1D-CNN can capture the combined features, by which the complex interactive relationship between different brain regions can be better represented.

Traditional CNN layers recognize local patterns through convolution, but high-order representations model complex feature interactions in deeper layers, capturing abstract patterns beyond local feature detection. Deeper layers learn abstract and composite features by combining lower-level elements, leading to more discriminative representations \cite{cnnhigh3}. High-order pooling operations capture correlations and dependencies that conventional linear operations in CNN may miss, enriching feature representations that are particularly important in medical image analysis, where complex patterns and subtle changes are crucial for accurate diagnosis \cite{cnnhigh2}.

For the upper triangular matrix of $C$, we denote a one-dimensional feature vector $x\in\mathbb{R}^k$, where $k=N(N-1)/2$. The convolution operation is performed by a one-dimensional convolution kernel to extract local features. Assume the convolution kernel size is $n$, the output feature map can be expressed as:
\begin{equation}
    y_i = \sum_{j=0}^{n-1}w_jx_{i+j}+b,
\end{equation}
where $w_j$ is the weight of the convolution kernel and $b$ is the bias term.

The MLP outputs the features $Z_{fc}\in\mathbb{R}^d$, where the transformation of each layer can be expressed as:
\begin{equation}
    z^{(l+1)} = \sigma(W^{(l)}z^{(l)}+b^{(l)}),
\end{equation}
where $W^{(l)}$ and $b^{(l)}$ are the weights and biases of the $l$th layer, and $\sigma$ is the activation function.

For calculating high-order representation, we define the HOP operator as follow:
\begin{equation}
    Z_{HOP} = Z_{fc}^TZ_{fc},
\end{equation}
where $Z_{HOP}$ is a real symmetric matrix. Each element $Z_{HOP}(i,j)$ in the matrix can be expressed as: $Z_{HOP}(i,j) = \langle Z_{fc}(i,:), Z_{fc}(j,:)\rangle $, which reflect the interaction strength and covariation relationship between them. This operation captures pairwise multiplicative interactions between different feature dimensions \cite{cnnhigh1}, enabling the extraction of interactions and co-variations between functional connectivity patterns, and providing richer high-order statistical features beyond the original FC.

The expression of the final high-order features is:
\begin{equation}
    \tilde{Z}_{HOP} = Concat(Z_{fc}, f_{MLP}(Z_{HOP})),
\end{equation}
where $\tilde{Z}_{HOP}$ represents high-order fusion features.

\subsection{FFC Module}
In this module, the final discriminative features for NDD diagnosis are obtained through the fusion of high-order features $\tilde{Z}$, which can be expressed as:
\begin{equation}
    \tilde{Z} = Concat(\tilde{Z}_{GHOP}, \tilde{Z}_{HOP}),
\end{equation}

The predictor comprises a MLP followed by a softmax function to generate a probability vector indicating the presence or absence of the disease for each subject. The prediction equation is as follows:

\begin{equation}
    \hat{y}=\mathrm{Softmax}(\mathrm{MLP}(\tilde{Z})),
\end{equation}
where $MLP$ denotes a multi-layer perceptron that processes the concatenated features $\tilde{Z}$. The softmax function then converts the MLP output into a probability vector indicating the likelihood of the presence or absence of the disease. Cross-entropy loss is employed to train the entire model.

\begin{equation}
    \mathcal{L}=-\frac1N\sum_{i=1}^N\left[y_i\log(\hat{y}_i)+(1-y_i)\log(1-\hat{y}_i)\right]
\end{equation}
where $N$ is the number of subjects, ${y}_i$ is the true label for the $i$-th subject, and $\hat{y}_i$ is the predicted probability for the $i$-th subject.

\section{Experiments}
\subsection{Datasets and Preprocessing}
The Autism Brain Imaging Data Exchange I (ABIDE-I) was released in August 2012, and it is the first initiative of the ABIDE project, involving 17 international research sites sharing rs-fMRI, anatomical, and phenotypic data. ABIDE-I has been extensively used in research. It includes 1112 subjects, with 539 subjects of ASD and 573 of typical controls. The ages of subjects are within 7-64 years old. For more information about collection parameters and site distribution, see the web page \url{https://fcon_1000.projects.nitrc.org/indi/abide/abide_I.html}.

The Autism Brain Imaging Data Exchange II (ABIDE-II), supported by the National Institute of Mental Health, was established to build on ABIDE-I's success in aggregating MRI data across 19 sites. To meet the requirement of larger data samples, ABIDE-II has collected the data from 1114 subjects, with 521 of ASD and 593 health controls, aged 5-64 years. ABIDE-II characterizes both the complexity of the connectome and the heterogeneity of ASD with enhanced phenotypic details and associated symptoms. Besides, it also includes longitudinal data from 38 individuals at two time points.  For more information about collection parameters and site distribution, see the web page \url{https://fcon_1000.projects.nitrc.org/indi/abide/abide_II.html}.

The ADHD-200 is a publicly available multi-site neuroimaging dataset designed to facilitate the study of ADHD. The ADHD-200 is a collaboration of 8 international imaging sites that have aggregated neuroimaging data from 362 children and adolescents with ADHD and 585 typically developing controls. These 947 datasets are composed of T1 and rs-fMRI data along with phenotypic information. The ADHD-200 dataset records the details of the ADHD diagnostic types of each subject, including ADHD-I (inattentive), ADHD-C (combined) and ADHD-HI (hyperactive/impulsive). For more information about collection parameters and site distribution, see the web page \url{http://fcon_1000.projects.nitrc.org/indi/adhd200/}.
The number of subjects included in this study and their demographics are given in Table~\ref{tab1}.

The DPARSF is used to preprocess the rs-fMRI data from ABIDE-I and ABIDE-II datasets \cite{ref-journal44} and Athena\textsuperscript{2} \cite{athena} pipeline is used to preprocess the rs-fMRI data from ADHD-200 dataset. The preprocessing procedure includes skull stripping, slice timing correction, and motion correction to minimize artifacts. Additionally, nuisance covariates, including signals from white matter, cerebrospinal fluid, and head motion, were regressed out. Next, fMRI images were normalized to Montreal Neurological Institute (MNI) space and underwent spatial smoothing using a Gaussian kernel with a full-width at half-maximum (FWHM) of 6 $\times$ 6 $\times$ 6 mm\textsuperscript{3}. BOLD signals underwent further processing through band-pass filtering (0.01 $\leq$ f $\leq$ 0.1 Hz) to eliminate high-frequency noise unrelated to neural activity and low-frequency drift in MRI scans.

AAL1 and Brainnetome Altase templates are two important and widely used brain atlases for neuroimaging analysis. AAL1 partitions the brain into 116 regions, providing a standardized framework for studying brain activity, while the Brainnetome Atlas offers a finer parcellation into 246 regions based on functional and structural connectivity. These templates are used to extract BOLD time series from specific brain regions in fMRI studies. 

\begin{table}
 \caption{Demographic statistics of the datasets used in this work.\label{tab1}}
  \centering
  \begin{tabular}{lllll}
    \toprule
    \textbf{Dataset}	& \textbf{Subgroup}	&\textbf{Number} & \textbf{Gender(M/F)}  & \textbf{Age(mean ± std.)}\\
\midrule
\multirow[m]{2}{*}{ABIDE-I}
& ASD	& 403 & 349/54 & 17.07 ± 7.95 \\
& NC	& 468 & 378/90 & 16.84 ± 7.23 \\

\midrule
\multirow[m]{2}{*}{ABIDE-II}
& ASD	& 243 &  203/40 & 13.82 ± 8.43  \\
& NC	& 289 &  184/105 & 12.63 ± 7.41 \\

\midrule
\multirow[m]{2}{*}{ADHD-200}
& ADHD	& 218 & 178/39 & 11.56 ± 5.91 \\
& NC	& 364 & 198/166 & 12.42 ± 8.62 \\
    \bottomrule
  \end{tabular}
\end{table}

\subsection{Experimental Setting}

\begin{figure}
    \centering
\includegraphics[width=0.98\textwidth]{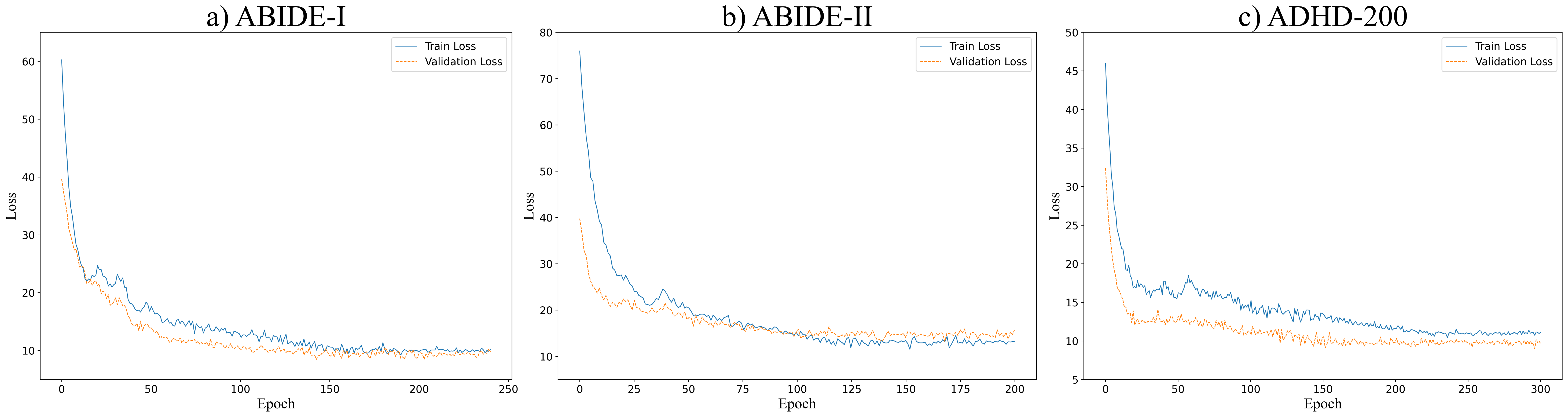}
\caption{{Training loss curves on three datasets. The plots show the convergence of the model during training: (a) ABIDE-I dataset with 240 epochs, (b) ABIDE-II dataset with 200 epochs, and (c) ADHD-200 dataset with 300 epochs.\label{loss}}}
\end{figure} 

The experiments were conducted by using a single server with NVIDIA RTX 3090 GPU. We developed our model using PyTorch, and every algorithm mentioned in this study is capable of running on a single GPU. Adaptive moment estimation (Adam) was employed for network optimization. For ABIDE-I (NC vs. ASD), the dropout rate is set to 0.3, the learning rate is set to 0.0001 and the maximum number of epochs is set to 240, the cutoff threshold is set to 19.03\%. For ABIDE-II (NC vs. ASD), the dropout rate is set to 0.25, the learning rate is set to 0.0001 and the maximum number of epochs is set to 200, the cutoff threshold is set to 17.14\%. For ADHD-200 (NC vs. ADHD), the dropout rate is set to 0.3, the learning rate is set to 0.0001 and the maximum number of epochs is set to 300, the cutoff threshold is set to 10.23\%. The results of classification are averaged over 10 times of cross validated test, where a train/validation/test ratio of 70\%/10\%/20\% was consistently applied across all comparison methods with balanced class distribution. {The training convergence curves of our model on three datasets are shown in Fig. \ref{loss}. The loss curves demonstrate stable and consistent convergence patterns across all datasets, with rapid initial descent in the early epochs followed by gradual stabilization, indicating effective learning and parameter optimization.} Five metrics of AUC, ACC, SEN, SPEC, and AVG are used to evaluate the classification performance.

\subsection{Competing Methods}
We conduct a comparative analysis of the proposed MHNet framework against twelve distinct traditional machine learning and deep learning methods.

One-dimensional feature of RV coefficient matrix or FC coefficient matrix obtained from different views of BFN are input to the traditional machine learning methods. In order to obtain the best classification performance, we experiment with varying the number of nodes in different hidden layers of the MLP and adjusting the penalty coefficients in the linear SVM. Additionally, we explore different k values (i.e., the number of neighbors) and select appropriate distance metrics (such as Euclidean or Manhattan distance) in KNN.

In addition to the traditional machine learning method, we list the comparative deep learning methods below:
\begin{enumerate}[1)]
\item 	{\bf BrainGNN\cite{ref-journal37}}: BrainGNN is an end-to-end graph neural network designed for fMRI analysis, capable of accurately identifying significant brain regions to decode task states and detect biomarkers. BrainGNN has been released at \url{https://github.com/xxlya/BrainGNN_Pytorch}.

\item	{\bf AL-NEGAT\cite{ref-journal45}}: AL-NEGAT (adversarial learning-based node-edge graph attention network) is a model designed to classify brain disease using both structural and functional MRI data. AL-NEGAT leverages both node and edge features through an attention-based mechanism and adversarial training methods to improve robustness and interpretability. AL-NEGAT has been released at \url{https://github.com/XiJiangLabUESTC/Node-Edge-Graph-Attention-Networks}. 

\item	{\bf MVS-GCN\cite{ref-journal38}}: MVS-GCN is a graph neural network guided by prior brain structure learning, which integrates graph structure learning with multi-task graph embedding and combines brain networks of varying sparsity levels as adjacency matrices for comprehensive feature representation. MVS-GCN has been released at \url{https://github.com/GuangqiWen/MVS-GCN}.

\item	{\bf Hi-GCN\cite{ref-journal36}}: Hi-GCN is an advanced neural network framework designed for learning graph embeddings from brain networks. It is specifically tailored to enhance accuracy of brain disorder prediction by considering both individual brain networks and the relationships between subjects in a population networks. Hi-GCN has been released at \url{https://github.com/haojiang1/hi-GCN}.

\item {\bf BrainNetCNN\cite{ref-journal19}}: BrainNetCNN is a convolutional neural network framework designed to predict clinical NDD based on brain networks. This method leverages novel convolutional filters that utilize the topological locality of brain networks. BrainNetCNN has been released at \url{https://github.com/nicofarr/brainnetcnnVis_pytorch}.

\item	{\bf BrainNetTF\cite{ref-journal46}}: BrainNetTF models brain networks as graphs with nodes of fixed size and order. This method proposes an readout operation that results in distinctive cluster-aware node embeddings and informative graph embeddings. BrainNetTF has been released at \url{https://github.com/Wayfear/BrainNetworkTransformer}.

\item   {\bf MAHGCN\cite{ref-journal48}}: MAHGCN is a framework for brain disorder diagnosis that utilizes multiscale brain atlases to construct hierarchical FCNs. It combines GCNs with atlas-guided pooling to extract and integrate multiscale topological features, significantly enhancing the accuracy of brain disorder predictions. MAHGCN has been released at \url{https://github.com/MianxinLiu/MAHGCN-code}.

\item	{\bf Com-BrainTF\cite{ref-conference47}}: Com-BrainTF is a transformer architecture for brain network analysis that incorporates community-specific information to enhance the accuracy and interpretability of fMRI data analysis. It features a hierarchical local-global transformer design that efficiently learns intra- and inter-community node embeddings. Com-BrainTF has been released at \url{https://github.com/ubc-tea/Com-BrainTF}.
\end{enumerate}

\subsection{Evaluation Metrics}
We evaluated the effectiveness of the MHNet framework by measuring five metrics: accuracy (ACC), sensitivity (SEN), specificity (SPEC), the area under the curve (AUC), and the average score (AVG) for each metric accordingly. In order to mitigate bias from a singular dataset split, we employed a 10-fold cross-validation approach during the evaluation phase.

To compare the performance of models, we use the almost stochastic order (ASO) test of statistical significance. For all tests, the significance level is set to $\alpha$ = 0.05 and adjusted using the Bonferroni correction when making multiple comparisons.

\section{Results}

\subsection{Classification Performance {and Model Efficiency}}
The classification results of using AAL1 atlas and Brainnetome atlas are shown in Table~\ref{tab2} and Table~\ref{tab3} respectively. Our proposed MHNet achieves the highest mean accuracy of 76.29\%, 75.16\%, 70.33\% on the datasets of ABIDE-I, ABIDE-II, and ADHD-200 respectively using the Brainnetome atlas, which is approximately 4\%–8\% higher than other state-of-the-art methods. The two ML methods show the worst performance. The original GCN exhibits the worst performance in the category of deep learning methods for ASD or ADHD classification.

{The proposed MHNet maintains a balanced trade-off between model complexity and computational efficiency. The input dimension varies across datasets due to different ROI definitions (e.g., 116 ROIs for AAL1 atlas). The network contains approximately 0.63M trainable parameters, with the HGNN module accounting for about 64\% of the total parameters due to its three parallel graph neural networks and the HCNN module accounting for about 36\% of the total parameters. During training on a single NVIDIA RTX 3090 GPU, the model requires around 16 GB of GPU memory. The inference training times on ABIDE-I, ABIDE-II, ADHD-200 are about 4.7 hours, 2.8 hours, 3.5 hours respectively.}

\begin{table}
	\caption{Classification performance for ASD (ASD vs. NC) on ABIDE I and ABIDE II, and for ADHD (ADHD vs. NC) on ADHD-200, using the AAL1 atlas. Results (shown as percentages) were derived from 10-fold cross-validation. Statistically significant differences from MHNet marked *. The best result in each category is highlighted in bold.\label{tab2}}
	\centering
	\begin{tabular}{lllllll}
		\toprule
		\textbf{Dataset} & \textbf{Method} & \textbf{AUC}  & \textbf{ACC} & \textbf{SEN}   & \textbf{SPEC}   & \textbf{AVG}        \\
		\midrule
		\multirow[m]{13}{*}{ABIDE-I} 
		& SVM & 61.46 ± 2.30* & 60.51 ± 2.81* & 60.60 ± 5.24* & 59.96 ± 1.90* & 60.13 ± 2.56*\\
		& KNN & 60.40 ± 3.05* & 58.68 ± 3.48* & 58.40 ± 1.51* & 60.49 ± 2.23* & 63.78 ± 2.12*\\
		& MLP & 62.06 ± 5.13* & 61.61 ± 4.54* & 62.93 ± 4.49* & 59.57 ± 3.07* & 63.44 ± 3.08*\\
		& GCN & 64.14 ± 5.39* & 63.49 ± 1.06* & 64.10 ± 1.35* & 60.73 ± 5.99* & 61.74 ± 0.28*\\
		& BrainGNN & 67.70 ± 2.15* & 65.70 ± 1.40* & 69.79 ± 0.51* & 65.73 ± 1.12* & 67.23 ± 2.53*\\
		& AL-NEGAT & 71.54 ± 2.58* & 69.33 ± 2.60* & 71.94 ± 1.11* & 68.63 ± 2.59* & 70.86 ± 4.22*\\
		& MVS-GCN & 69.30 ± 1.65* & 67.39 ± 3.36* & 67.64 ± 1.46* & 66.59 ± 4.41* & 67.44 ± 0.84*\\
		& Hi-GCN & 71.83 ± 2.41* & 68.60 ± 2.86* & 68.62 ± 1.28* & 65.84 ± 3.31* & 68.00 ± 2.30*\\
		& BrainNetCNN & 68.24 ± 3.37* & 69.10 ± 0.33* & 72.38 ± 1.96* & 69.67 ± 2.51 & 69.45 ± 2.20*\\
		& BrainNetTF & 71.13 ± 2.57* & 70.15 ± 2.21* & 71.09 ± 1.46* & 69.25 ± 1.99* & 71.54 ± 2.91*\\
		& MAHGCN & 72.56 ± 2.88* & 72.06 ± 2.53* & 72.77 ± 2.66* & 70.79 ± 0.85 & 71.44 ± 3.85\\
		& Com-BrainTF & 73.06 ± 1.45* & 71.81 ± 2.80* & 73.12 ± 0.72* & 69.14 ± 1.37* & 71.73 ± 3.91*\\
		& \textbf{MHNet (Ours)} & \textbf{76.31 ± 1.02} & \textbf{73.27 ± 2.14} & \textbf{76.83 ± 1.45} & \textbf{71.24 ± 2.39} & \textbf{74.41 ± 2.06}\\
		
		\midrule
		\multirow[m]{13}{*}{ABIDE-II} 
		& SVM & 60.59 ± 1.67* & 59.62 ± 1.29* & 61.23 ± 4.80* & 58.54 ± 3.24* & 60.00 ± 4.39*\\
		& KNN & 61.81 ± 1.75* & 60.38 ± 2.51* & 63.29 ± 2.83* & 59.63 ± 5.61* & 62.54 ± 3.12*\\
		& MLP & 63.35 ± 3.27* & 60.27 ± 3.65* & 61.27 ± 1.59* & 59.03 ± 4.36* & 61.84 ± 2.98*\\
		& GCN & 63.96 ± 2.53* & 63.22 ± 2.39* & 64.54 ± 2.06* & 60.13 ± 2.59* & 63.27 ± 3.90*\\
		& BrainGNN & 68.70 ± 1.09* & 68.35 ± 1.40* & 69.57 ± 0.51* & 67.73 ± 1.12* & 69.03 ± 3.53*\\
		& AL-NEGAT & 71.63 ± 1.26* & 70.87 ± 3.03* & 71.03 ± 2.43* & 68.54 ± 1.48* & 70.32 ± 3.22*\\
		& MVS-GCN & 70.52 ± 1.65* & 68.69 ± 2.16* & 69.34 ± 2.76* & 67.18 ± 2.41* & 69.35 ± 2.39*\\
		& Hi-GCN & 70.04 ± 2.41* & 70.37 ± 1.23* & 71.24 ± 2.56* & 68.23 ± 2.01* & 71.24 ± 2.30*\\
		& BrainNetCNN & 69.24 ± 1.29* & 68.10 ± 1.73* & 71.38 ± 1.96* & 67.67 ± 1.35* & 70.45 ± 2.20*\\
		& BrainNetTF & 72.13 ± 2.57* & 69.15 ± 2.21* & 71.09 ± 1.46* & 67.25 ± 1.99* & 69.54 ± 2.91*\\
		& MAHGCN & 71.39 ± 1.46 & 71.06 ± 0.94* & 70.77 ± 1.42* & \textbf{70.47 ± 0.97} & 71.64 ± 1.39*\\
		& Com-BrainTF & 71.89 ± 0.75* & 70.81 ± 1.57* & 72.02 ± 0.94* & 69.14 ± 1.62* & 70.39 ± 3.36*\\
		& \textbf{MHNet (Ours)} & \textbf{73.94 ± 1.57} & \textbf{72.83 ± 0.94} & \textbf{73.61 ± 0.82} & 70.23 ± 1.79 & \textbf{72.00 ± 3.58}\\

		\midrule
		\multirow[m]{13}{*}{ADHD-200} 
		& SVM & 60.60 ± 2.88* & 59.76 ± 2.68* & 59.61 ± 4.14* & 59.34 ± 2.50* & 59.02 ± 2.77*\\
		& KNN & 58.46 ± 2.62* & 58.01 ± 2.23* & 57.98 ± 1.34* & 59.73 ± 1.34* & 60.81 ± 1.09*\\
		& MLP & 60.99 ± 4.04* & 60.65 ± 4.21* & 62.10 ± 2.96* & 58.93 ± 2.65* & 62.20 ± 3.19*\\
		& GCN & 63.46 ± 2.62* & 65.78 ± 1.09* & 63.61 ± 1.88* & 62.72 ± 5.25* & 64.29 ± 0.95*\\
		& BrainGNN & 64.07 ± 2.81* & 67.16 ± 1.50* & 61.93 ± 0.49* & 60.73 ± 1.51* & 62.34 ± 2.15*\\
		& AL-NEGAT & 69.33 ± 3.56 & 68.60 ± 2.91* & 65.27 ± 1.97* & 63.14 ± 2.52* & 66.88 ± 3.52\\
		& MVS-GCN & 67.49 ± 2.91* & 66.72 ± 3.05* & 68.80 ± 1.60* & 65.11 ± 4.13* & 67.53 ± 1.11*\\
		& Hi-GCN & 68.84 ± 2.15* & 67.39 ± 2.76* & 69.42 ± 2.74* & 65.39 ± 3.33* & 67.15 ± 3.90*\\
		& BrainNetCNN & 68.33 ± 2.32* & 66.71 ± 1.30* & 64.81 ± 1.48* & 64.01 ± 1.41 & 66.38 ± 2.90*\\
		& BrainNetTF & 69.26 ± 3.09* & 67.64 ± 2.39* & 69.53 ± 0.83* & 67.70 ± 2.92* & 68.06 ± 1.53*\\
		& MAHGCN & 68.34 ± 3.56 & 68.13 ± 1.88 & 68.22 ± 2.67* & 67.91 ± 1.06* & 68.67 ± 3.34*\\
		& Com-BrainTF & 70.10 ± 2.65* & 67.73 ± 2.45* & 68.22 ± 0.96* & \textbf{69.59 ± 0.73} & 69.23 ± 3.75*\\
		& \textbf{MHNet (Ours)} & \textbf{71.12 ± 2.71} & \textbf{69.64 ± 1.39} & \textbf{71.57 ± 0.93} & 68.25 ± 1.27 & \textbf{70.17 ± 1.49}\\
		
		\bottomrule
	\end{tabular}
\end{table}

\begin{table}
	\caption{Classification performance for ASD (ASD vs. NC) on ABIDE I and ABIDE II, and for ADHD (ADHD vs. NC) on ADHD-200, using the Brainnetome atlas. Results (shown as percentages) were derived from 10-fold cross-validation. Statistically significant differences from MHNet marked *. The best result in each category is highlighted in bold.} \label{tab3}
	\centering
	\begin{tabular}{lllllll}
		\toprule
		\textbf{Dataset} & \textbf{Method} & \textbf{AUC}  & \textbf{ACC} & \textbf{SEN}   & \textbf{SPEC}   & \textbf{AVG}        \\
		\midrule
		\multirow[m]{13}{*}{ABIDE-I} 
		& SVM & 63.08 ± 3.19* & 62.37 ± 3.82* & 62.30 ± 4.83* & 61.57 ± 2.84* & 61.57 ± 3.35*\\
		& KNN & 61.16 ± 3.39* & 60.14 ± 2.65* & 60.51 ± 2.56* & 62.71 ± 1.35* & 63.16 ± 1.17*\\
		& MLP & 64.03 ± 3.89* & 63.20 ± 4.62* & 64.58 ± 3.25* & 61.73 ± 2.80* & 64.37 ± 2.46*\\
		& GCN & 65.10 ± 6.59* & 64.35 ± 1.62* & 66.13 ± 2.34* & 59.68 ± 6.52* & 63.82 ± 1.27*\\
		& BrainGNN & 69.94 ± 2.62* & 68.68 ± 1.62* & 66.28 ± 0.62* & 65.68 ± 0.62* & 66.23 ± 2.36*\\
		& AL-NEGAT & 71.29 ± 3.19* & 70.37 ± 2.54* & 71.13 ± 2.17* & 68.21 ± 2.39* & 70.29 ± 3.10*\\
		& MVS-GCN & 72.51 ± 2.12* & 71.33 ± 2.92* & 72.13 ± 2.42* & 69.24 ± 3.85* & 71.30 ± 1.39*\\
		& Hi-GCN & 74.55 ± 3.23* & 72.31 ± 2.64* & 70.34 ± 2.53* & 68.03 ± 2.45* & 71.81 ± 3.59*\\
		& BrainNetCNN & 72.59 ± 2.91* & 71.56 ± 0.93* & 72.21 ± 1.35* & 67.59 ± 1.31* & 72.24 ± 2.85\\
		& BrainNetTF & 74.93 ± 2.37* & 73.28 ± 1.97* & 72.11 ± 1.27* & 68.53 ± 2.71* & 70.89 ± 1.73*\\
		& MAHGCN & 73.21 ± 3.02* & 73.52 ± 1.63* & 73.02 ± 2.14* & 71.55 ± 1.38 & 72.57 ± 3.03\\
		& Com-BrainTF & 74.12 ± 1.72* & 73.58 ± 1.81* & 74.31 ± 0.51* & 70.64 ± 0.38* & 74.91 ± 3.41\\
		& \textbf{MHNet (Ours)} & \textbf{78.68 ± 2.07} & \textbf{76.29 ± 1.39} & \textbf{76.43 ± 1.21} & \textbf{72.65 ± 2.02} & \textbf{76.01 ± 2.57}\\
		
		\midrule
		\multirow[m]{13}{*}{ABIDE-II} 
		& SVM & 61.43 ± 2.23* & 60.15 ± 2.37* & 61.25 ± 3.74* & 59.44 ± 1.06* & 60.57 ± 4.25*\\
		& KNN & 62.37 ± 1.09* & 61.24 ± 1.76* & 63.83 ± 1.67* & 60.82 ± 2.15* & 62.07 ± 2.37*\\
		& MLP & 62.83 ± 2.42* & 62.02 ± 3.00* & 63.15 ± 2.61* & 60.54 ± 2.41* & 63.15 ± 1.98*\\
		& GCN & 63.80 ± 2.76* & 62.85 ± 2.03* & 64.96 ± 3.62* & 58.17 ± 3.87* & 62.65 ± 1.84*\\
		& BrainGNN & 69.37 ± 3.27* & 66.34 ± 2.35* & 67.71 ± 2.49* & 65.28 ± 3.74* & 67.77 ± 2.75*\\
		& AL-NEGAT & 70.14 ± 0.80* & 68.11 ± 0.57* & 70.95 ± 3.41* & 68.95 ± 3.22* & 69.11 ± 3.55*\\
		& MVS-GCN & 69.24 ± 3.93* & 67.01 ± 2.30* & 69.97 ± 2.12* & 67.80 ± 3.23* & 68.92 ± 3.91*\\
		& Hi-GCN & 70.30 ± 2.74* & 69.94 ± 1.00* & 70.17 ± 3.81* & 68.64 ± 2.33* & 69.86 ± 1.95*\\
		& BrainNetCNN & 69.32 ± 1.43* & 67.13 ± 3.21* & 68.02 ± 2.10* & 65.14 ± 2.49* & 66.98 ± 0.57\\
		& BrainNetTF & 71.42 ± 2.66* & 70.89 ± 2.64* & 71.97 ± 1.66* & 69.03 ± 2.80* & 71.76 ± 2.89*\\
		& MAHGCN & 72.17 ± 1.76* & 71.27 ± 2.03* & 72.56 ± 2.94* & 69.89 ± 0.71 & 71.33 ± 2.83\\
		& Com-BrainTF & 72.73 ± 2.85* & 71.18 ± 1.24* & 72.92 ± 0.95* & 69.41 ± 1.60* & 71.35 ± 1.77\\
		& \textbf{MHNet (Ours)} & \textbf{76.28 ± 1.82} & \textbf{75.16 ± 1.63} & \textbf{74.39 ± 1.57} & \textbf{71.33 ± 1.45} & \textbf{74.29 ± 2.02}\\

		\midrule
		\multirow[m]{13}{*}{ADHD-200} 
		& SVM & 61.15 ± 2.21* & 60.20 ± 3.00* & 60.30 ± 4.31* & 59.65 ± 2.04* & 59.83 ± 3.14*\\
		& KNN & 59.09 ± 2.71* & 58.38 ± 2.42* & 58.09 ± 1.61* & 60.18 ± 0.78* & 61.47 ± 0.70*\\
		& MLP & 61.75 ± 3.55* & 61.30 ± 3.82* & 62.63 ± 2.74* & 59.26 ± 2.19* & 63.14 ± 2.34*\\
		& GCN & 60.84 ± 5.93* & 63.19 ± 0.97* & 63.79 ± 1.68* & 57.42 ± 5.57* & 61.43 ± 3.56*\\
		& BrainGNN & 64.66 ± 2.62* & 62.78 ± 1.62* & 62.33 ± 0.62* & 61.21 ± 0.62* & 63.00 ± 2.36*\\
		& AL-NEGAT & 64.03 ± 3.19* & 63.21 ± 2.54* & 62.79 ± 2.17* & 63.95 ± 2.39* & 63.90 ± 3.10*\\
		& MVS-GCN & 66.96 ± 2.12* & 64.04 ± 2.92* & 66.33 ± 2.42* & 63.98 ± 3.85* & 65.05 ± 1.39*\\
		& Hi-GCN & 67.49 ± 3.23* & 66.97 ± 2.64* & 65.84 ± 2.53* & 64.14 ± 2.45* & 66.08 ± 3.59*\\
		& BrainNetCNN & 69.08 ± 2.91* & 68.31 ± 0.93* & 69.42 ± 1.35* & 65.64 ± 1.31* & 68.08 ± 2.85\\
		& BrainNetTF & 68.90 ± 2.37* & 68.26 ± 1.97* & 68.29 ± 1.27* & 66.03 ± 2.71* & 67.37 ± 1.73*\\
		& MAHGCN & 69.56 ± 3.02* & 67.76 ± 1.63* & 68.88 ± 2.14* & 68.33 ± 1.38 & 68.47 ± 3.03\\
		& Com-BrainTF & 70.12 ± 1.72* & 68.14 ± 1.81* & 70.99 ± 0.51* & 67.05 ± 0.38* & 69.25 ± 3.41\\
		& \textbf{MHNet (Ours)} & \textbf{72.02 ± 0.86} & \textbf{70.33 ± 0.76} & \textbf{71.14 ± 1.81} & \textbf{68.84 ± 1.34} & \textbf{70.75 ± 2.68}\\
		
		\bottomrule
	\end{tabular}
\end{table}

\subsection{Ablation Study}
\subsubsection{Influence of Cutoff Threshold}
\begin{figure}
    \centering
\includegraphics[width=1\textwidth]{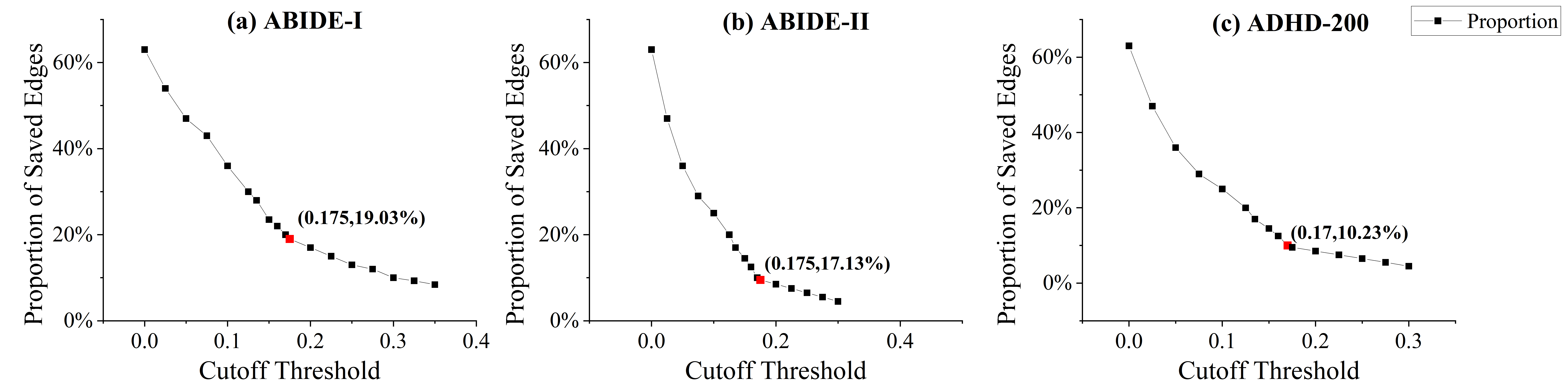}
\caption{Illustration of the relationship curve between the percentage of retained edges and the cutoff threshold.\label{fig2}}
\end{figure}

For HGNN, we use the obtained RV coefficient matrix to construct the adjacency matrix of the brain hierarchical network of each data set, but the RV coefficient matrix is not a sparse binary matrix. In order to achieve the best classification effect, threshold processing is required for the matrix. Specifically, if the RV coefficient between brain area $i$ and brain area $j$ is less than the cutoff threshold $\gamma$, the value at $(i, j)$ of the RV coefficient matrix is set to 0. As shown in Figure~\ref{fig2}, in order to select a suitable cutoff threshold, we plotted the relationship curve between the percentage of retained edges and the cutoff threshold, and selected the inflection point of the curve as the cutoff point. The cutoff threshold at this point is used to process the RV coefficient matrix into a sparse binary adjacency matrix.

\begin{figure}
    \centering
\includegraphics[width=1\textwidth]{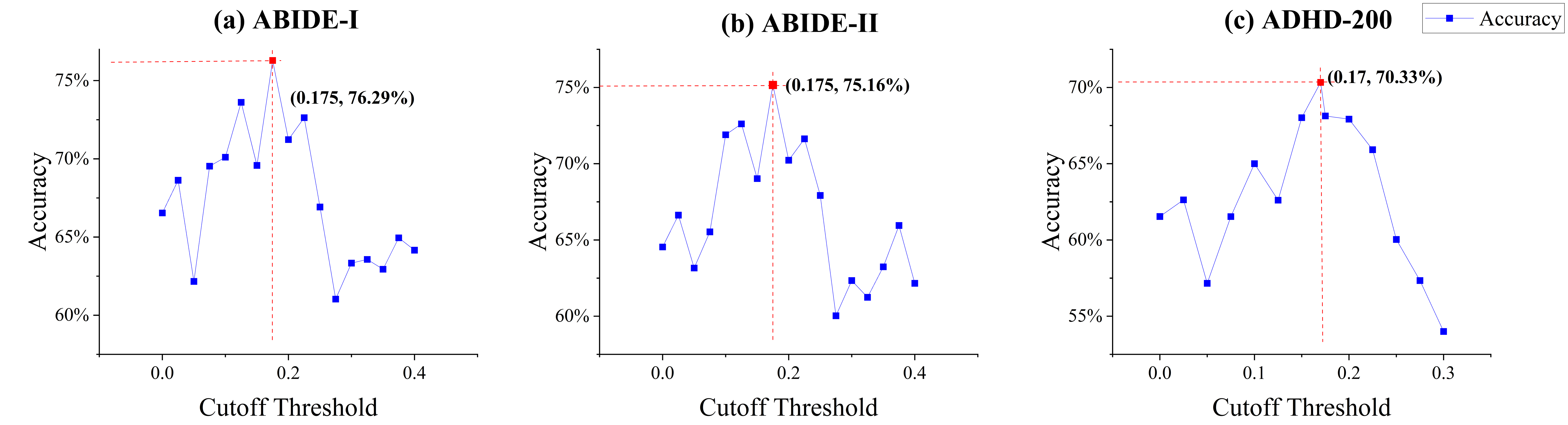}
\caption{Illustration of classification performance achievable with different cutoff thresholds.\label{fig3}}
\end{figure}

Figure~\ref{fig3} shows the classification effect that MHNet can achieve when taking different cutoff thresholds. We can see that the best classification effect is achieved when the inflection point of the relationship curve is selected.

\subsubsection{Influence of Brain Atlas Selection}
\begin{figure}
    \centering
\includegraphics[width=1\textwidth]{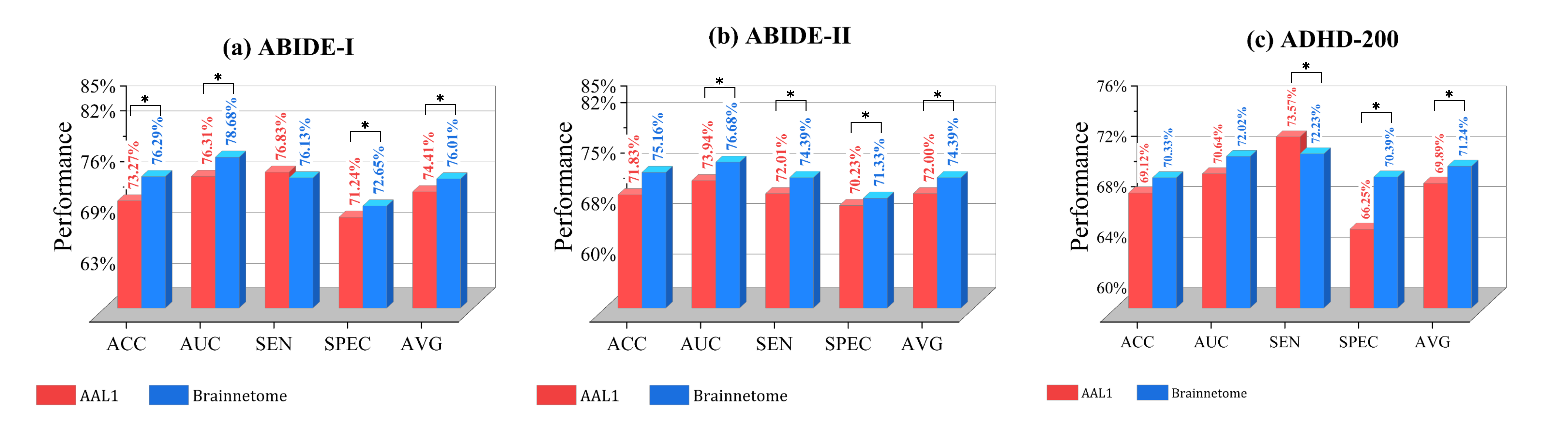}
\caption{Comparision of the classification performance between using AAL1 and Brainnetome. * indicates a statistically significant differences.\label{fig4}}
\end{figure}

In order to study the robustness of MHNet with respect to selecting different brain atlas, we use the atlases of AAL1 and Brainnetome to construct graphs with different numbers of nodes. The classification results are shown in Figure~\ref{fig4}. Our proposed MHNet model is built based on the hierarchical structure of BFN. The finer the  hierarchical structure is, the richer the brain FC and spatial information that the model can utilize. As we can see in Figure~\ref{fig4}, Brainnetome atlas show superior classification performance than AAL1. This is because that higher resolution and more detailed brain parcellation are conducive to building the brain hierarchical structure with more sophisticated information. In addition, Brainnetome atlas derived from multimodal data takes FC inforamtion into account, so that the parcellation is in line with the actual BFN.

\subsubsection{Influence of GNN Encoder}
\begin{figure}
	\centering
	\includegraphics[width=1\textwidth]{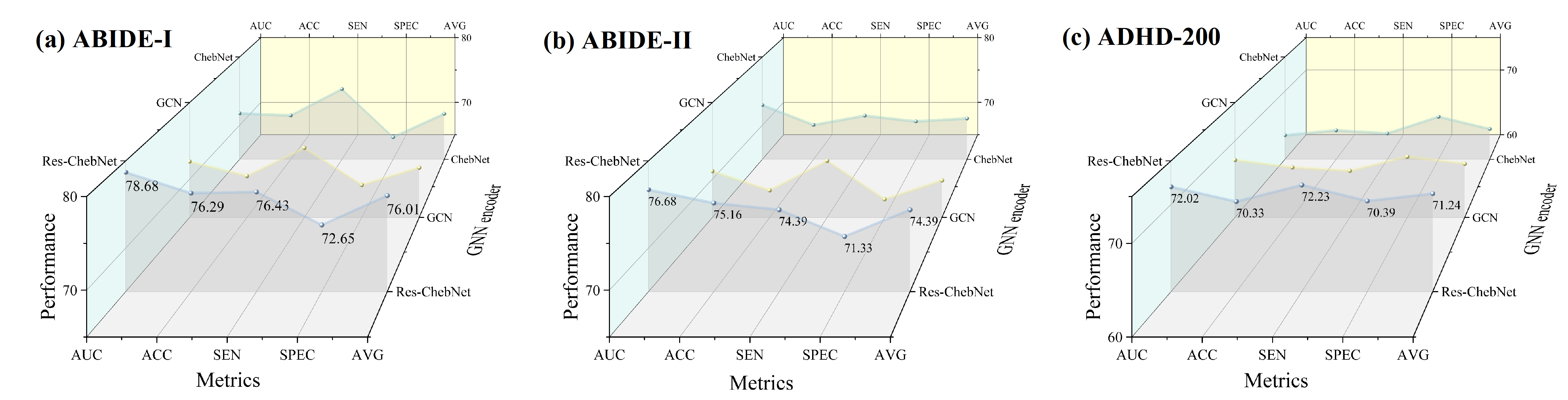}
	\caption{The best classification performance that the MHNet can achieve using GNN Encoders of Res-ChebNet, GCN, and ChebNet respectively.\label{fig5}}
\end{figure}

To study the impact of different GNN encoders on the classification performance of MHNet, we replaced the Res-ChebNet blocks with GCN and ChebNet, and used Brainnetome atlas to evaluate the performance on three datasets. As shown in Figure~\ref{fig5}, Res-ChebNet outperforms GCN and ChebNet on all three datasets. This means that Res-ChebNet has higher efficiency and accuracy in capturing and processing complex BFNs.

\subsubsection{Influence of Multi-view and High-order Feature Extractors}
In order to verify the effectiveness of our proposed MHNet based on multi-view and high-order feature representation, we conducted ablation experiments on the multi-view and high-order feature representation constructed by the brain hierarchical network.

\begin{table}[!ht]
	\caption{Classification performance for ASD (ASD vs. NC) classification results on ABIDE I and ABIDE II, and ADHD (ADHD vs. NC) classification results on ADHD-200 using varied multi-view data and advanced features. Percentages represent the outcomes from 10-fold cross-validation. Statistically significant differences from MHNet marked *. The best result in each category is highlighted in bold.\label{tab4}}
	\begin{tabular}{llll}
		\toprule
		\textbf{Dataset}	& \textbf{Method}  & \textbf{ACC}    & \textbf{AUC}\\
		\midrule
		\multirow[m]{8}{*}{ABIDE-I} 
		& GNN(Only Brain-LAN)\textsuperscript{1}	& 63.39 ± 2.59*	& 65.15 ± 1.53*\\
		& GNN\textsuperscript{2}	& 66.62 ± 0.76*	& 67.37 ± 2.54*\\
		& CNN\textsuperscript{3}	& 64.39 ± 1.85*	& 65.93 ± 1.36*\\
		& HGNN\textsuperscript{4}	& 71.48 ± 1.23*	& 73.35 ± 1.46*\\
		& HCNN\textsuperscript{5}	& 70.19 ± 2.04*	& 72.66 ± 1.37*\\
		& HCNN + GNN\textsuperscript{6}	& 71.93 ± 2.29* & 73.02 ± 4.32\\
		& HGNN + CNN\textsuperscript{7}	& 73.06 ± 2.03 & 74.58 ± 1.03*\\
		& HGNN + HCNN\textsuperscript{8}	& \textbf{76.29 ± 1.39}	& \textbf{78.68 ± 2.07}\\
		
		\midrule
		\multirow[m]{8}{*}{ABIDE-II} 
		& GNN(Only Brain-LAN)\textsuperscript{1}	& 64.45 ± 1.03*	& 66.15 ± 1.53*\\
		& GNN\textsuperscript{2}	& 65.31 ± 2.03*	& 67.15 ± 1.74*\\
		& CNN\textsuperscript{3}	& 65.91 ± 1.54*	& 65.19 ± 1.49*\\
		& HGNN\textsuperscript{4}	& 69.49 ± 1.71*	& 71.23 ± 0.99*\\
		& HCNN\textsuperscript{5}	& 68.45 ± 3.13*	& 70.22 ± 1.95*\\
		& HCNN + GNN\textsuperscript{6}	& 70.59 ± 1.62* & 73.28 ± 1.57*\\
		& HGNN + CNN\textsuperscript{7}	& 71.35 ± 1.43* & 73.53 ± 2.23\\
		& HGNN + HCNN\textsuperscript{8}	& \textbf{75.16 ± 1.63}	& \textbf{76.28 ± 1.82}\\
		
		\midrule
		\multirow[m]{8}{*}{ADHD-200} 
		& GNN(Only Brain-LAN)\textsuperscript{1}	& 62.18 ± 2.96*	& 65.29 ± 1.37*\\
		& GNN\textsuperscript{2}	& 62.03 ± 2.23*	& 63.48 ± 1.27*\\
		& CNN\textsuperscript{3}	& 63.26 ± 1.59*	& 64.03 ± 1.93*\\
		& HGNN\textsuperscript{4}	& 67.12 ± 0.74*	& 68.23 ± 1.25*\\
		& HCNN\textsuperscript{5}	& 67.38 ± 2.19*	& 67.69 ± 1.73*\\
		& HCNN + GNN\textsuperscript{6}	& 68.05 ± 1.21* & 69.59 ± 2.24*\\
		& HGNN + CNN\textsuperscript{7}	& 68.54 ± 2.37 & 70.34 ± 1.56*\\
		& HGNN + HCNN\textsuperscript{8}	& \textbf{70.33 ± 0.76}	& \textbf{72.02 ± 0.86}\\
		
		\bottomrule
	\end{tabular}
	\footnotetext[1]{GNN (Only Brain-LAN) means that only the Brian-LAN layer in G-IBHN-G is used to obtain brain FC features and high-order features and HCNN branches are not used.}
	\footnotetext[2]{GNN means that a complete BFN with hierarchical structure in G-IBHN-G is used, but high-order features and HCNN branches are not used.}
	\footnotetext[3]{CNN means that a complete BFN with hierarchical structure in G-IBHN-G is used, but high-order features and HCNN branches are not used.}
	\footnotetext[4]{HGNN uses high-order feature representation on the basis of GNN.}
	\footnotetext[5]{HCNN uses high-order feature representation on the basis of CNN.}
	\footnotetext[6]{HCNN + GNN means that the FC information of the brain in Euclidean space is also used on the basis of HGNN, but high-order features are not used.}
	\footnotetext[7]{HGNN + CNN means that the FC information of the brain in Euclidean space is also used on the basis of HGNN, but high-order features are not used.}
	\footnotetext[8]{HGNN + HCNN is the complete MHNet model.}
\end{table}


As shown in Table~\ref{tab4}, without encoding high-order features and multi-view information, GNN (Only Brain-LAN) demonstrates basic performance, and all evaluation indicators are relatively low. Compared with GNN (Only Brain-LAN), the performance of GNN is improved, which means that the hierarchical structure of BFN can capture more useful features. Similarly, CNN achieves comparable results, suggesting that Euclidean-based FC features also contribute to the classification performance. Based on GNN, HGNN is proposed to encode high-order features, which significantly improves the performance of the model, especially in capturing the complex FC of the brain. HCNN, which introduces high-order features on the basis of CNN, also shows notable performance enhancement. Combining HCNN with GNN yields better results than using GNN alone, highlighting the benefit of high-order features. HCNN + GNN combines Euclidean high-order features with non-Euclidean network structure, showing moderate improvement compared to single-view methods. Apart from HGNN, we also add CNN which encodes the FC information of the brain in Euclidean space, resulting in a HGNN + CNN model. Although this model introduces the Euclidean information in FC, the performance improvement is not as significant as that of HGNN. The high-order features may play a more important role in capturing complex activity patterns of the brain. HGNN + HCNN shows the best performance with respect to all evaluation indicators, indicating the effectiveness of introducing the complementary multi-view high-order features.

\subsubsection{{Influence of Network Architecture Parameters}}

\begin{table}[!t]
\caption{Different Network Architecture Parameters on Classification Performance (\%). Statistically significant differences from MHNet with optimal parameters marked *. The best results are shown in \textbf{bold}.}
\label{cheb}
\centering
\begin{tabular}{ccccccccc}
\hline
\multirow{2}{*}{{Parameter}} & \multirow{2}{*}{{Value}} & \multicolumn{2}{c}{{ABIDE-I}} & \multicolumn{2}{c}{{ABIDE-II}} & \multicolumn{2}{c}{{ADHD-200}} \\
\hhline{~~------}
& & {ACC} & {AUC} & {ACC} & {AUC} & {ACC} & {AUC} \\
\hline
\multirow{5}{*}{{ChebNet Order (K)}} 
& {1} & {64.45*} & {65.89*} & {63.23*} & {64.98*} & {59.12*} & {61.89*} \\
& {2} & {72.78*} & {73.12*} & {70.45*} & {72.87*} & {67.91*} & {69.45*} \\
& {3} & \textbf{{76.29}} & \textbf{{78.68}} & \textbf{{75.16}} & \textbf{{76.28}} & \textbf{{70.33}} & \textbf{{72.02}} \\
& {4} & {73.89*} & {74.12*} & {72.56*} & {73.91*} & {69.17*} & {70.12*} \\
& {5} & {71.12*} & {71.67*} & {69.89*} & {70.45*} & {65.12*} & {68.78*} \\
\hline
\multirow{4}{*}{{\#ChebConv Modules}} 
& {1} & {68.12*} & {69.11*} & {67.37*} & {68.03*} & {63.43*} & {64.72*} \\
& {2} & {73.15*} & {73.98*} & {72.32*} & {73.87*} & {67.23*} & {69.29*} \\
& {3} & \textbf{{76.29}} & \textbf{{78.68}} & \textbf{{75.16}} & \textbf{{76.28}} & \textbf{{70.33}} & \textbf{{72.02}} \\
& {4} & {74.38*} & {75.15*} & {73.52*} & {73.73*} & {68.28*} & {70.25*} \\
\hline
\end{tabular}
\end{table}

{To investigate the impact of key architectural parameters on model performance, we conducted extensive ablation studies focusing on the polynomial order (K) of ChebNet and the number of ChebConv modules in Brainnetome Atlas. For the polynomial order, we varied K from 1 to 5 while keeping other parameters fixed. As shown in Table \ref{cheb}, the model achieves optimal performance with K=3 on all three datasets, with accuracies of 76.29\%, 75.16\%, and 70.33\% for ABIDE-I, ABIDE-II, and ADHD-200 respectively. When K $>$ 3, the performance begins to degrade due to increased model complexity and potential overfitting. Similarly, we examined the effect of ChebConv module depth by varying the number of layers from 1 to 4. The results in Table \ref{cheb} demonstrate that three ChebConv modules provide the best balance between model capacity and computational efficiency. Adding more layers leads to marginal improvements while significantly increasing computational cost. These findings guided our final architecture design, which uses K=3 and three ChebConv modules.}

\section{Discussion}
\subsection{Interpretability Analysis}
\begin{figure}
    \centering
\includegraphics[width=1\textwidth]{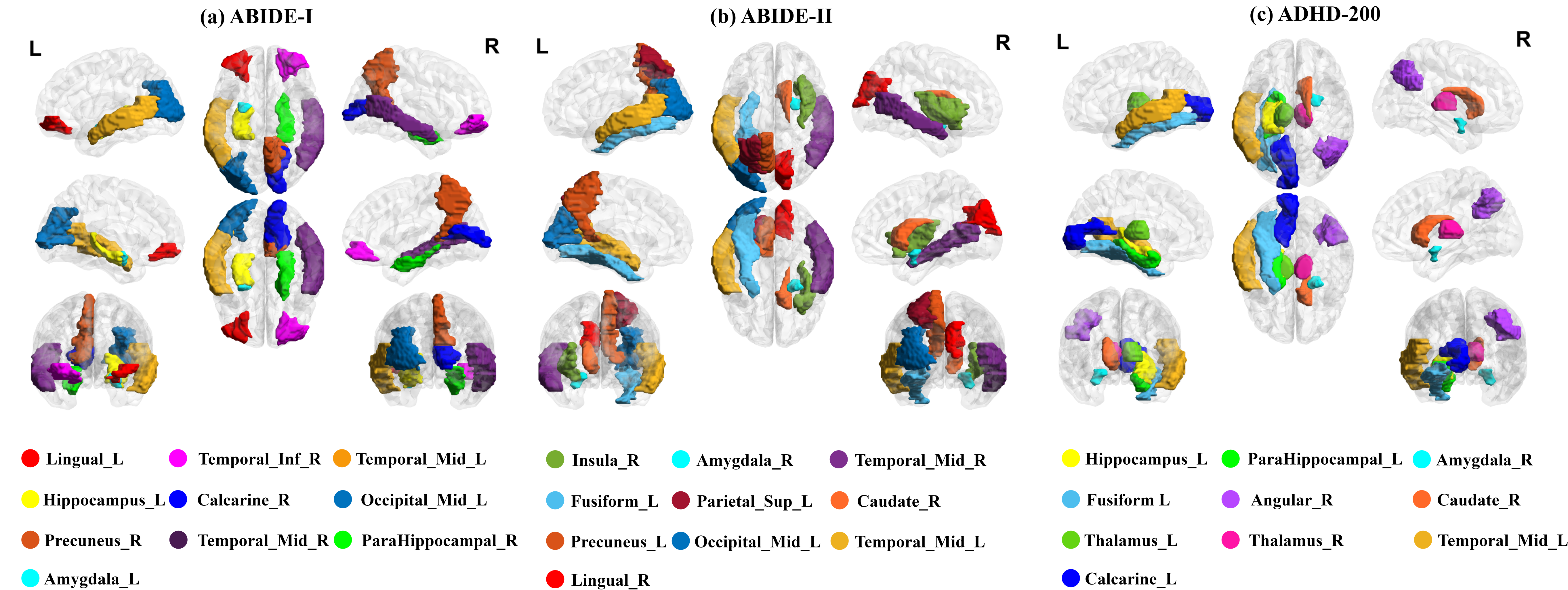}
\caption{Top 10 important brain regions associated with ASD/ADHD based on the datasets of ABIDE-I, ABIDE-II, and ADHD-200.\label{fig6}}
\end{figure}

We first discuss the important brain regions associated with NDD when using MHNet. By integrating HGNN and HCNN, the brain regions most relevant to NDD can be effectively identified. HGNN module encodes both node features and connectivity, by which the most representative brain regions and connections related to the NDD in the classification tasks can be identified. In HCNN module, the local brain region features with strongest response to NDD classification are identified by analyzing the convolutional layer activation map. The top 10 identified brain regions which are most relevant to the NDD are visualized in Figure \ref{fig6} for experimental verification on ABIDE-I, ABIDE-II, and ADHD-200 datasets.

ASD involves abnormalities in multiple brain regions, including the lingual gyrus, emporal inferior gyrus, hippocampus, amygdala, calcarine fissure, and occipital middle gyrus \cite{{ref-journal49}}. The lingual gyrus plays an important role in visual processing, especially in facial recognition and emotional understanding. The abnormality of lingual gyrus may lead to difficulties in visual information integration and social interaction for patients with ASD \cite{ref-journal50}. The inferior temporal gyrus and the temporal middle gyrus play an important role in language processing and social cognition, and their functional abnormalities may explain the language comprehension and episodic memory impairments for patients with ASD \cite{ref-journal51}. While abnormalities in amygdala are associated with the problems in emotional processing and social behavior regulation \cite{ref-journal52}. Abnormalities in the calcarine fissure and the occipital middle gyrus may affect primary and higher-order processing of visual information, which explains the challenges that ASD patients face in processing visual social cues \cite{ref-journal53}. In addition, the insula, crucial for emotion processing and interoception, along with precuneus, fusiform gyrus, parietal superior lobule, and caudate nucleus show significant abnormalities in ASD patients \cite{ref-journal54}. The precuneus governs self-reflection and social cognition, fusiform gyrus controls facial recognition, and superior parietal lobule manages spatial cognition, all showing abnormalities impacting ASD patients' social and cognitive functions. The caudate nucleus plays an important role in motor control, learning, and executive function, and its functional abnormalities may be related to repetitive behaviors, narrow interests, and executive dysfunction in ASD patients \cite{ref-journal55}. The combined changes in these brain regions lead to a wide range of challenges in cognitive, emotional, and social functions for ASD patients.

ADHD involves abnormalities in multiple brain regions, including the left hippocampus and left parahippocampal gyrus, which play an important role in memory formation and emotion regulation, and their abnormalities may lead to deficits in working memory and episodic memory \cite{ref-journal56}. The right amygdala plays an important role in emotion processing and response control, and its functional abnormalities may lead to emotional overreaction and impulsive behavior \cite{ref-journal57}. Abnormalities in the left calcarine fissure may affect visual attention and information processing. The right caudate nucleus and bilateral thalamus play an important role in motor control and attention regulation, and their abnormalities may lead to excessive movement, impulsive behavior and difficulty in attention regulation \cite{ref-journal58}. Abnormalities in the left middle temporal gyrus may affect language comprehension and social cognition \cite{ref-journal59}. The combined abnormal changes in these areas together explain the wide range of cognitive, emotional and behavioral disorders for ADHD patients.

\subsection{Brain Feature Analysis}
The combination of multi-view features of node and connectivity can provide more comprehensive information of BFN in MHNet framework. The FC matrix is estimated by calculating the the similarity of neural activity of different brain regions. CNN was used to encode Euclidean space features of FC matrix in this work. The hierarchical structure of BFN is modeled in non-Euclidean space. By encoding the graph nodes features and edges, GNN can more naturally capture the complex topological characteristics of the BFN \cite{network}. The combination of CNN and GNN can not only capture the node and connectivity information of global BFN, but also characterize the local hierarchical structure of BFN. 

Multi-view high-order features delineate the intricate and abstract patterns of BFN by integrating complementary features from both Euclidean and non-Euclidean spaces, thereby enhance the robustness of MHNet and improve the classification accuracy.

\begin{figure}
	\centering
	\includegraphics[width=1\textwidth]{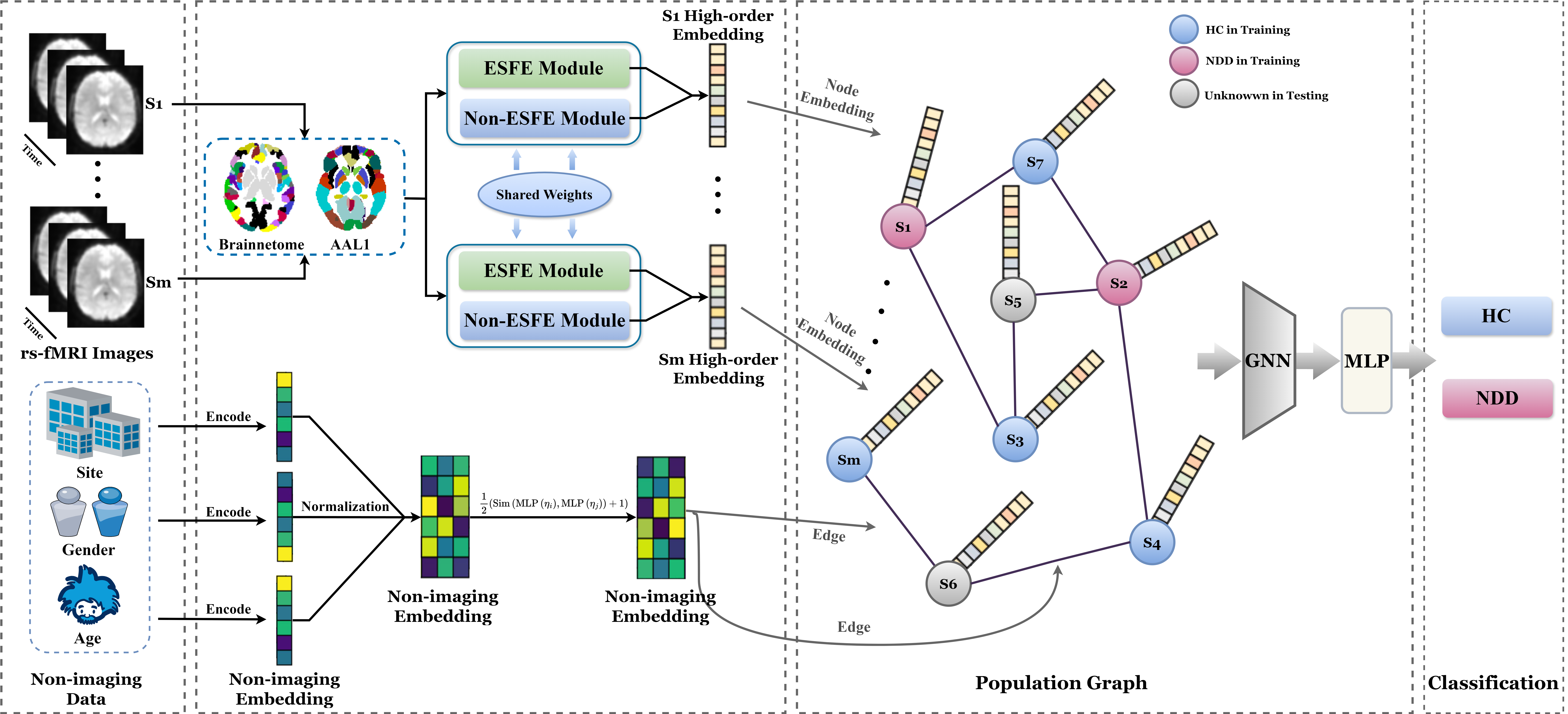}
	\caption{Architecture of population graph incorporating phenotypic information.}\label{fig7}
\end{figure}

\subsection{Non-imaging Data}

In this work, our MHNet model achieves significant diagnostic results. When designing MHNet, we focused on designing models that are good at identifying and analyzing local brain regions and biomarkers associated with diseases. However, our proposed MHNet model does not take into accout the non-imaging data. Non-imaging data such as clinical symptoms and behavioral assessments are likely to play an important role in the diagnosis of NDD \cite{ref-journal60}.

To further explore the impact of non-imaging data, we considered the relationships among subjects and tried to exploit phenotypic information such as gender, age, and site. First, leveraging MHNet's exceptional performance, we employ it to generate precise embeddings that effectively represent brain networks. Then, we construct a population graph for subject classification, with each node representing an individual subject's brain and each edge capturing the relationship between pairs of subjects. The specific framework of the population graph is shown in Figure \ref{fig7}.

To achieve a comprehensive representation of all subjects $S=\{{S}_{1},{S}_{2},\ldots,{S}_{m}\}$ using both imaging and non-imaging data, we construct a population graph $G_{\text{p}} = \{ {V}', {A}' \}$. Here, $m$ denotes the number of subjects, so the set $V'$ contains $m$ nodes. Each node in $V'$ corresponds to an embedding $Y$ of a subject $S$, obtained by reshaping the output of the MHNet into a vector form. The adjacency matrix $A'$ for the graph is defined as ${A}^{\prime}={C}*{W}$, where $C$ is a binarized connectivity matrix derived from both imaging and non-imaging data. To construct $C$, we first create a similarity matrix ${M}_{1}$ using the embedding features $Y$:

\begin{equation}
{M}_1=\exp\left(-\frac{[\rho({Y}_i,{Y}_j)]^2}{2\sigma^2}\right),
\end{equation}
where $\rho$ represents the correlation distance between embeddings, and $\sigma$ is the mean of $[\rho({Y}_i,{Y}_j)]^2$. Next, we develop a node similarity metric ${M}_2$ by incorporating phenotypic information (e.g., gender) to obtain more useful clinical auxiliary information. These two matrices are then combined using the Hadamard product to form ${C}^{\prime}={M}_1*{M}_2$.

\begin{table}[!ht]
	\caption{Classification performance for ASD (ASD vs. NC) classification results on ABIDE I and ABIDE II, and ADHD (ADHD vs. NC) classification results on ADHD-200 using non-imaging data. Percentages represent the outcomes from 10-fold cross-validation. Statistically significant differences from MHNet marked *. The best result in each category is highlighted in bold.}\label{population}
	\begin{tabular}{ccccc}
		\toprule
		\textbf{Dataset}	& \textbf{Method}  & \textbf{Non-imaging Data} & \textbf{ACC}    & \textbf{AUC}\\
		\midrule
		\multirow[m]{4}{*}{ABIDE-I}
		& GCN (Regional Brain Graph)\textsuperscript{1} & $\times$	& 66.62 ± 0.76*	& 67.37 ± 2.54*\\
		& GCN (Population Graph)\textsuperscript{2} & $\checkmark$	& 69.13 ± 1.24*	& 69.31 ± 2.38*\\
		& MHNet (Regional Brain Graph)\textsuperscript{3} &  $\times$	& 76.29 ± 1.39*	& \textbf{78.68 ± 2.07}\\
		& \textbf{MHNet (Population Graph)}\textsuperscript{4} & $\checkmark$	& \textbf{77.03 ± 2.47}	& 78.51 ± 1.39\\
		
		\midrule
		\multirow[m]{4}{*}{ABIDE-II} 
		& GCN (Regional Brain Graph)\textsuperscript{1} & $\times$	& 65.31 ± 2.03*	&  67.15 ± 1.74*\\
		& GCN (Population Graph)\textsuperscript{2} & $\checkmark$	& 66.29 ± 1.53*	& 68.93 ± 1.89*\\
		& MHNet (Regional Brain Graph)\textsuperscript{3} &  $\times$	& 75.16 ± 1.63	& 76.28 ± 1.82\\
		& \textbf{MHNet (Population Graph)}\textsuperscript{4} & $\checkmark$	& \textbf{75.73 ± 2.02}	& \textbf{77.39 ± 1.31}\\
		
		\midrule
		\multirow[m]{4}{*}{ADHD-200}
		& GCN (Regional Brain Graph)\textsuperscript{1} & $\times$	& 62.03 ± 2.23*	&  63.48 ± 1.27*\\
		& GCN (Population Graph)\textsuperscript{2} & $\checkmark$	&  63.19 ± 3.04*	& 63.29 ± 2.69*\\
		& MHNet (Regional Brain Graph)\textsuperscript{3} &  $\times$	& 70.33 ± 0.76*	&  72.02 ± 0.86*\\
		& \textbf{MHNet (Population Graph)}\textsuperscript{4} & $\checkmark$	& \textbf{72.12 ± 2.17}	& \textbf{72.39 ± 1.53}\\
		
		\bottomrule
	\end{tabular}
	\footnotetext[1]{GCN (Regional Brain Graph): GCN for regional brain graph classification, without phenotypic data.}
	\footnotetext[2]{GCN (Population Graph): GCN with phenotypic data for population graph node classification.}
	\footnotetext[3]{MHNet (Regional Brain Graph): GCN for regional brain graph classification, without phenotypic data.}
	\footnotetext[4]{MHNet (Population Graph): MHNet to capture node features and construct population graph with phenotypic data.}
\end{table}

From $C'$, we obtain the binarized connection graph $C$. The sparse weight matrix $W$ is constructed by calculating each element ${W}_{ij}$ based on the normalized non-imaging data ${\eta}_{i}$ and ${\eta}_{j}$ of two subjects:
\begin{equation}
{W}_{ij}=\frac{Sim\left(\text{MLP}(\eta_i),\text{MLP}(\eta_j)\right)+1}{2},
\end{equation}

In this equation, the MLP layers share weights, and $Sim$ denotes the cosine similarity between the outputs of the MLPs applied to ${\eta}_{i}$ and ${\eta}_{j}$.

Next, we use GCN and MLP to classify nodes and get the classification result $y$. The specific operations are as follows:
\begin{equation}
	y=MLP(GCN(Y,A^{\prime})).
\end{equation}

As shown in Table \ref{population}, incorporating non-imaging data through population graphs yields modest improvements in classification performance. While the MHNet with population graph achieves the best performance across all datasets, the improvements over MHNet with regional brain graph are relatively small. This limited improvement can be attributed to our initial focus in MHNet design, which primarily emphasized the identification and analysis of local brain regions and disease-related biomarkers rather than the integration of non-imaging features. Nevertheless, statistical analysis reveals that MHNet-based methods (both with and without population graphs) significantly outperform GNN-based approaches, demonstrating the robustness of MHNet's brain network feature extraction capabilities. The results suggest that while non-imaging data provides complementary information, the core strength of MHNet lies in its ability to effectively capture disease-specific patterns from brain imaging data.

\subsection{Study Limitations}
While MHNet demonstrates promising results in NDD diagnosis, several aspects warrant further investigation. Although validated on three public datasets using AAL1 and Brainnetome Atlas templates, incorporating more diverse datasets and brain atlas templates would enhance generalizability across different clinical settings and demographics. The computational complexity increases with multiple views and hierarchical levels, necessitating optimization for real-time clinical applications, while more effective multi-view integration strategies could improve clinical utility. Beyond basic demographics, expanding phenotypic features to include behavioral assessments, cognitive profiles, and environmental factors, coupled with systematic evaluation of their relative contributions to diagnostic accuracy, could provide more nuanced NDD characterization and guide clinical data collection priorities.

\section{Conclusion}
The MHNet offers a significant advancement in diagnosing NDD, such as ASD and ADHD using rs-fMRI data. By integrating both GNN and CNN methodologies, MHNet effectively captures hierarchical and high-order feature representations. The inclusion of residual ChebNet in the HGNN module improves gradient flow, enhances feature propagation, and increases model flexibility. The multi-view feature integration and extraction of hierarchical structure of BFNs characterize both global and local topological information of BFN in non-Euclidean and Euclidean space. 

Our study addresses the limitations of current deep learning models used for diagnosing NDD with rs-fMRI in non-uniform data adaptability and capturing high-level features. Encoding hierarchical structure of BFN and integrating high-order non-Euclidean space and Euclidean features, MHNet has shown superior performance to SOTA on three NDD datasets, demonstrating its powerful feature extraction and classification capabilities. For future research directions, we aim to generalize our framework to diagnose additional brain disorders and explore the integration of MHNet with non-image information for brain disorder diagnosis.

\section*{Declarations}
\bmhead{Funding}
This research was supported by the National Natural Science Foundation of China with grant number 31870979 and the Hong Kong Polytechnic University Start-up Fund (Project ID: P0053210).
\bmhead{Conflict of interest}
The authors declare no conflicts of interest.
\bmhead{Ethics approval and consent to participate}
Informed consent was obtained from all subjects involved in the study.
\bmhead{Data availability}
The used data in this study from ADHD-200 \url{(https://fcon_1000.projects.nitrc.org/indi/adhd200/)}, ABIDE-I \url{(https://fcon_1000.projects.nitrc.org/indi/abide/abide_I)} and ABIDE-II \url{(https://fcon_1000.projects.nitrc.org/indi/abide/abide_II)} is available publicly.

\bmhead{Author contribution}
Conceptualization: Yueyang Li, Weiming Zeng, NiZhuan Wang. 
Methodology: Yueyang Li, Weiming Zeng, NiZhuan Wang.  
Software: Yueyang Li, Wenhao Dong, Lei Wang. 
Validation: Yueyang Li, Weiming Zeng, NiZhuan Wang, Luhui Cai, Hongjie Yan, Lingbin Bian.
Formal analysis:, Yueyang Li, Luhui Cai, Wenhao Dong. 
Investigation: Yueyang Li, Weiming Zeng, NiZhuan Wang, Hongjie Yan, Lingbin Bian, Wenhao Dong, Lei Wang. 
Data curation: Yueyang Li, Wenhao Dong, Luhui Cai. 
Writing—original draft preparation: all authors. 
Writing—review and editing: all authors. 
Visualization: Yueyang Li, Weiming Zeng, NiZhuan Wang, Wenhao Dong, Luhui Cai.
Supervision, Weiming Zeng, NiZhuan Wang. 
Project administration: Weiming Zeng, NiZhuan Wang. 
Funding acquisition: Weiming Zeng. 
All authors have read and agreed to the published version of the manuscript.

\bibliography{sn-bibliography}

\end{document}